\documentclass[preprint,12pt]{elsarticle}

\usepackage{amssymb}
\usepackage{amsmath}
\usepackage{graphicx}
\usepackage{subcaption}
\usepackage{xcolor}
\definecolor{verde}{rgb}{0.0, 0.6, 0.0}
\definecolor{rojo}{rgb}{0.96, 0.76, 0.76}
\usepackage{adjustbox}
\usepackage{booktabs}
\usepackage{makecell}
\usepackage{multirow}
\usepackage{array}
\usepackage{comment}
\usepackage{algorithmic}
\usepackage[acronym]{glossaries}
\usepackage{algorithm}
\usepackage{orcidlink}
\usepackage{url}
\usepackage{comment}

\newcolumntype{C}[1]{>{\centering\arraybackslash}p{#1}}

\usepackage{lineno}

\newif\ifanonymous
\anonymousfalse

\journal{Engineering Applications of Artificial Intelligence}

\newacronym{ISR}{ISR}{Intelligence, Surveillance and Reconnaissance}
\newacronym{GEOINT}{GEOINT}{Geospatial Intelligence}
\newacronym{ADS-B}{ADS-B}{Automatic Dependent Surveillance -- Broadcast}
\newacronym{GNSS}{GNSS}{Global Navigation Satellite System}
\newacronym{PSR}{PSR}{Primary Surveillance Radar}
\newacronym{SSR}{SSR}{Secondary Surveillance Radar}
\newacronym{VHF}{VHF}{Very High Frequency}
\newacronym{UHF}{UHF}{Ultra High Frequency}
\newacronym{LoS}{LoS}{Line of Sight}
\newacronym{UAV}{UAV}{Unmanned Aerial Vehicle}
\newacronym[firstplural={Unmanned Aerial Vehicles (UAVs)}]{UAVs}{UAVs}{Unmanned Aerial Vehicles}

\newacronym{CubeSat}{CubeSat}{Cube Satellite}
\newacronym{LEO}{LEO}{Low Earth Orbit}
\newacronym{SSO}{SSO}{Sun-Synchronous Orbit}
\newacronym{ESA}{ESA}{European Space Agency}
\newacronym{NASA}{NASA}{National Aeronautics and Space Administration}
\newacronym{SWAP}{SWAP}{Size, Weight and Power}
\newacronym{ADCS}{ADCS}{Attitude Determination and Control System}
\newacronym{GSD}{GSD}{Ground Sample Distance}
\newacronym{TT&C}{TT\&C}{Telemetry, Tracking and Command}
\newacronym{OBDP}{OBDP}{On-Board Data Processing}
\newacronym{COTS}{COTS}{Commercial Off-The-Shelf}
\newacronym{SEE}{SEE}{Single Event Effect}
\newacronym{TID}{TID}{Total Ionising Dose}
\newacronym{SRAM}{SRAM}{Static Random Access Memory}

\newacronym{AI}{AI}{Artificial Intelligence}
\newacronym{DL}{DL}{Deep Learning}
\newacronym{CNN}{CNN}{Convolutional Neural Network}
\newacronym[firstplural={Convolutional Neural Networks (CNNs)}]{CNNs}{CNNs}{Convolutional Neural Networks}
\newacronym{R-CNN}{R-CNN}{Region-based Convolutional Neural Network}
\newacronym{RPN}{RPN}{Region Proposal Network}
\newacronym{HOG}{HOG}{Histogram of Oriented Gradients}
\newacronym{SVM}{SVM}{Support Vector Machine}
\newacronym{NMS}{NMS}{Non-Maximum Suppression}
\newacronym{IoU}{IoU}{Intersection over Union}
\newacronym{GIoU}{GIoU}{Generalised Intersection over Union}
\newacronym{mAP}{mAP}{mean Average Precision}
\newacronym{FPS}{FPS}{Frames Per Second}
\newacronym{FLOPs}{FLOPs}{Floating-point Operations}
\newacronym{GFLOPs}{GFLOPs}{Giga Floating-point Operations}
\newacronym{YOLO}{YOLO}{You Only Look Once}
\newacronym{SSD}{SSD}{Single Shot MultiBox Detector}
\newacronym{DETR}{DETR}{DEtection TRansformer}
\newacronym{RT-DETR}{RT-DETR}{Real-Time Detection TRansformer}
\newacronym{ViT}{ViT}{Vision Transformer}
\newacronym{CLIP}{CLIP}{Contrastive Language--Image Pre-training}
\newacronym{VAE}{VAE}{Variational Autoencoder}
\newacronym{UNET}{UNET}{U-shaped Network}
\newacronym{SAHI}{SAHI}{Slicing Aided Hyper Inference}
\newacronym{DOTA}{DOTA}{Dataset for Object Detection in Aerial Images}

\newacronym{LDM}{LDM}{Latent Diffusion Model}
\newacronym[firstplural={Latent Diffusion Models (LDMs)}]{LDMs}{LDMs}{Latent Diffusion Models}
\newacronym{GAN}{GAN}{Generative Adversarial Network}
\newacronym[firstplural={Generative Adversarial Networks (GANs)}]{GANs}{GANs}{Generative Adversarial Networks}
\newacronym{FLUX}{FLUX}{Flow Matching Text-to-Image Model}
\newacronym{LoRA}{LoRA}{Low-Rank Adaptation}
\newacronym{LLM}{LLM}{Large Language Model}
\newacronym{GGUF}{GGUF}{GPT-Generated Unified Format}

\newacronym{ASIC}{ASIC}{Application-Specific Integrated Circuit}
\newacronym{ALU}{ALU}{Arithmetic Logic Unit}
\newacronym{CPU}{CPU}{Central Processing Unit}
\newacronym{CUDA}{CUDA}{Compute Unified Device Architecture}
\newacronym{DRAM}{DRAM}{Dynamic Random Access Memory}
\newacronym{RAM}{RAM}{Random Access Memory}
\newacronym{VRAM}{VRAM}{Video Random Access Memory}
\newacronym{TPU}{TPU}{Tensor Processing Unit}
\newacronym{VPU}{VPU}{Vision Processing Unit}
\newacronym{FPGA}{FPGA}{Field-Programmable Gate Array}
\newacronym{INT8}{INT8}{8-bit Integer Quantisation}
\newacronym{AMP}{AMP}{Automatic Mixed Precision}
\newacronym{TOPS}{TOPS}{Tera Operations Per Second}
\newacronym{GPU}{GPU}{Graphics Processing Unit}

\newacronym{SAR}{SAR}{Synthetic Aperture Radar}
\newacronym{IR}{IR}{Infrared}
\newacronym{Backbone}{Backbone}{Feature-extraction backbone}
\newacronym{PANet}{PANet}{Path Aggregation Network}
\newacronym{C3K2}{C3K2}{Cross Stage Partial Bottleneck with 3 convolutions}
\newacronym{C2f}{C2f}{Cross Stage Partial Bottleneck with 2 convolutions}
\newacronym{C2PSA}{C2PSA}{Cross-Stage Partial with Spatial Attention}
\newacronym{PSA}{PSA}{Pointwise Spatial Attention}
\newacronym{SPPF}{SPPF}{Spatial Pyramid Pooling -- Fast}
\newacronym{COCO}{COCO}{Common Objects in Context}
\makenoidxglossaries

\graphicspath{{figs/}}

\begin{document}

\begin{frontmatter}

\title{Towards Autonomous Aircraft Surveillance from Nanosatellites through On-Board Inference and Generative Data Augmentation}

\ifanonymous
\else
\cortext[cor1]{Corresponding author}

\author[A]{Antonio Delgado-Rosa \orcidlink{0009-0000-8099-7442}}
\ead{antonio.delgado3@alu.uclm.es}

\author[A]{David Muñoz-Valero \orcidlink{0000-0002-2509-9911}}
\ead{david.munoz@uclm.es}

\author[B]{Enrique Adrian Villarrubia-Martin \orcidlink{0009-0002-8006-5711}} \ead{enrique.villarrubia@uclm.es}

\author[A]{Juan Moreno-Garcia \orcidlink{0000-0003-2430-145X}\corref{cor1}}
\ead{juan.moreno@uclm.es}

\address[A]{Escuela de Ingeniería Industrial y Aeroespacial de Toledo, Department of Technologies and Information Systems, Universidad de Castilla--La Mancha, Avenida Carlos III, s/n, Toledo, 45071, Spain}

\address[B]{Escuela Superior de Informática, Department of Technologies and Information Systems, Universidad de Castilla-La Mancha, Paseo de la Universidad 4, Ciudad Real, 13071, Spain}
\fi

\begin{abstract}
Airborne surveillance from low Earth orbit is hindered by two interconnected bottlenecks: nanosatellites have a limited downlink budget, yet the conventional approach still transmits terabytes of raw imagery to the ground for processing, and open satellite datasets for aircraft are scarce and severely class-imbalanced. These limitations either delay timely decision-making or prevent standard detectors from learning robust representations of rare aircraft classes. In this paper, a workflow that combines on-board inference with generative data augmentation is proposed to address both limitations jointly. Inference is executed on a 6U CubeSat equipped with a low-power edge tensor accelerator, while a diffusion model fine-tuned through low-rank adaptation generates synthetic minority-class imagery. This synthetic output is automatically annotated, pseudo-labelled, by an intermediate detector and merged with classically augmented samples. The results show that the balanced dataset increases global mean average precision from 77.9\% to 82.2\%, with the minority class rising from F1~=~0.683 to F1~=~0.811, and that the quantised detector fits the on-chip memory and projects 25--30 frames per second on orbit. This approach contrasts with the conventional bent-pipe architecture, in which the satellite acts as a passive data collector. Therefore, the computational tests support the proposed workflow as a decision-support tool for real-time, autonomous airborne surveillance from nanosatellites.
\end{abstract}


\end{frontmatter}



\section{Introduction}\label{sec:introduction}

Continuous monitoring of the airspace and the Earth's surface is essential for safety, emergency management and air-traffic control. Traditionally, these surveillance tasks have relied on ground-based infrastructure, such as primary and secondary radar networks, or on the manual review of aerial imagery. However, these approaches share well-known limitations: incomplete geographical coverage, high dependence on expensive installations, and response times that may delay critical decision-making~\cite{skolnik1980introduction,strohmeier2014realities}. Over the last decade, the irruption of nanosatellites has democratised orbital observation. Their reduced size and manufacturing cost have enabled the deployment of constellations that offer a high spatial resolution together with a revisit frequency never reached before~\cite{sweeting2018modern,poghosyan2017cubesat}. Despite these operational advantages, the massive acquisition capacity of on-orbit sensors has given rise to a new technological bottleneck: the volume of imagery generated far exceeds the capacity of traditional radio-frequency links to download it to the ground~\cite{furano2020towards}. The challenge is to process this immense quantity of visual telemetry efficiently, in order to detect relevant targets without requiring constant human intervention.

This work is motivated by two complementary gaps that, to the best of the authors' knowledge, have not yet been addressed jointly in the literature. The first gap concerns the architectural paradigm of current Earth-observation missions, which still follows a centralised conception inherited from the early space era, commonly denoted as \textit{bent pipe}. Under this scheme, the satellite acts as a mere relay in space: the entire payload of raw imagery is downlinked, and the analysis is delegated to high-performance ground servers. While these servers are able to execute dense predictive models, they relegate advanced orbital systems to the passive role of data collectors. As a result, on-board subsystems lack the situational consistency required to autonomously deduce which fragments of the gathered telemetry possess real operational value. The second gap concerns the training data: open satellite datasets for aircraft are scarce and severely imbalanced. In the standard \textit{HRPlanesV2} \cite{hrplanesv2_dataset} dataset, military aircraft outnumber the minority helicopter class by a factor of four, which prevents standard one-stage detectors from learning robust representations of rare classes. The state of the art usually addresses these two gaps in isolation: either algorithms are run on the ground with the latency penalty of the downlink, or theoretical models are trained on orbit-ready data that suffocate minority classes, or synthetic images are produced without considering the thermal and power constraints of space hardware. This paper unifies all these dimensions into a single coherent workflow.

To overcome the first gap, the workflow executes inference on board a CubeSat of the 6U form factor equipped with a Google Coral Edge Tensor Processing Unit (TPU) as the On-Board Data Processing (OBDP) accelerator. The detector is constrained to fit the 8\,MB of on-die Static Random-Access Memory (SRAM) available on the accelerator after INT8 quantisation, an explicit Size, Weight, and Power (SWaP)-aware  design condition that drives both the architecture selection and the compression stage. To overcome the second gap, a generative data-augmentation pipeline based on FLUX~\cite{blackforest2024flux} and Low-Rank Adaptation (LoRA)~\cite{hu2021loralowrankadaptationlarge} is proposed: a custom LoRA token is trained on the minority class, the model is quantised in GGUF format to keep the fine-tuning within reach of consumer hardware, and the synthesised images are pseudo-labelled by an intermediate detector and merged with classically augmented samples. Classical augmentation alone is shown to be insufficient for the helicopter class (F1 $=$ 0.683), whereas the addition of synthetic data lifts it to F1 $=$ 0.811, approaching the performance of the majority classes.

The main contributions of this paper are the following:
\begin{itemize}
  \item \textbf{A hardware-first methodology for space-grade AI deployment.} The physical constraints of the target platform, namely the 6U CubeSat bus, the Google Coral Edge TPU, and the 8\,MB SRAM budget, are established first, and only then is a detector architecture selected and adapted to fit within them.

  \item \textbf{A generative data-augmentation pipeline targeted at the minority class.} A custom FLUX+LoRA workflow produces synthetic helicopter imagery in four operational environments, which is afterwards pseudo-labelled by an intermediate YOLO detector and merged with classically augmented samples to yield a balanced dataset.

  \item \textbf{A SWaP-aware benchmark of one-stage detectors for space-grade hardware.} Three architectures (SSD MobileNet V3, YOLO11n and RT-DETR-L) are compared under the same unbalanced aerial dataset and against the same 8\,MB SRAM budget of the Coral Edge TPU. YOLO11n emerges as the best precision--latency compromise and is kept as reference.

  \item \textbf{An on-board concept of operations on a 6U CubeSat.} The selected detector is exported to INT8, mapped to the Edge TPU SRAM, and integrated with Slicing Aided Hyper Inference (SAHI) evaluation of 1024$\times$1024 patches, achieving 82.2\% mean Average Precision (mAP@50) with a projected on-orbit throughput of 25--30 Frames Per Second (FPS).
  \ifanonymous
\else
  The source code, FLUX LoRA adapters,
  ComfyUI workflows and trained weights are released in an open repository (\url{https://github.com/Antonio23013/TFG-DETECCION-DE-OBJETOS}) to ensure that all experiments can be fully reproduced and expanded upon by the community.
  \fi
\end{itemize}

The rest of the paper is structured as follows: Section~\ref{sec:related_work} reviews recent works showing the progress made in object detection in aerial imagery, generative data augmentation and edge computing for space. Due to the scarcity and strict physical limitations of space-grade components, this work adopts a hardware-first methodology: the physical capabilities must be established before an artificial intelligence model can be feasibly selected. Consequently, Section~\ref{sec:platform} describes the mission and platform constraints of the 6U CubeSat, including the optical payload, the Edge TPU accelerator, and the communications subsystem. Following this hardware definition, Section~\ref{sec:methods} presents the materials and methods, namely the \textit{HRPlanesV2} dataset, the data-augmentation pipeline, and the candidate architectures evaluated against those physical constraints. Section~\ref{sec:results} describes the experiments and results obtained, showing the performance of the selected architecture in the unbalanced, classically augmented, and synthetically balanced scenarios, and analysing the deployable INT8 model on orbit. Section~\ref{sec:discussion} discusses the practical 
applicability of the proposal in Intelligence, Surveillance, and Reconnaissance (ISR) and Geospatial Intelligence (GEOINT) scenarios, while acknowledging its limitations. Finally, Section~\ref{sec:conclusions} sets out conclusions and further work derived from this paper.
\section{Related Work}\label{sec:related_work}

This section reviews the literature most relevant to the objective presented in this work. The review is organised around the three technological threads that the proposed workflow integrates: edge computing for space missions, generative data augmentation, and object detection in aerial imagery.

\subsection{Edge computing for space}\label{sec:rw_edge}

The state of the art in satellite observation reveals a significant technological gap in the communications architecture. The prevailing operational paradigm follows a centralised conception inherited from the early space era, denoted as \textit{bent pipe}, in which the satellite acts as a mere relay and the entire raw payload is
downlinked for ground processing. Under this scheme, on-board subsystems lack the situational awareness required to autonomously discard redundant telemetry, which perpetuates the downlink bottleneck discussed in Section~\ref{sec:introduction}. Moving the \gls{OBDP} on board transforms the platform from a passive sensor into an autonomous node, but it demands accelerators compatible with the \gls{SWAP} budget of a \gls{CubeSat} and, at the same time, resilient to the radiation environment of \gls{LEO}. Two reference missions have pioneered this transition: the European Space Agency \gls{ESA} $\varphi$-sat-1, which demonstrated on-board cloud screening with a Myriad 2 vision processor, and the \gls{ESA} CISERES mission~\cite{esa_ciseres_2024}, led by Deimos, which applies on-board \gls{AI} for civil-security applications. A growing body of work has characterised the suitability of commercial off-the-shelf accelerators for such missions, including Intel Movidius Myriad, the Google Coral Edge \gls{TPU} and the NVIDIA Jetson family~\cite{cass2019taking,prokscha2023efficient,chanoui2025trends}, as well as their behaviour under radiation and single-event effects~\cite{furano2020towards,mystkowska2025hardware,lentaris2023performance}. In the same line, Magalh\~aes et al.~\cite{magalhaes2023benchmarking} benchmarked several edge devices for single-stage object detection in this journal, reporting that the trade-off between precision, latency and power is highly device-dependent. The present work adopts the same benchmarking philosophy but targets a different domain and an additional constraint: the selected detector must fit the on-die \gls{SRAM} of the accelerator after \gls{INT8} quantisation, so that the whole inference path can run without external memory and within the power envelope of a 6U \gls{CubeSat}.

\subsection{Generative data augmentation}\label{sec:rw_gan}

Deep-learning detectors remain strongly dependent on the quality and volume of training data. In the aerospace domain, manually collecting and annotating tens of thousands of captures of specific aircraft --- such as military helicopters in varied operational environments --- is logistically unfeasible, which motivates the use of synthetic data.
\glspl{GAN}~\cite{goodfellow2014generativeadversarialnetworks} were the first generative family applied to this end, with variants such as CycleGAN used to simulate infrastructure or to alter meteorological conditions in aerial photographs. However, \glspl{GAN} suffer from training instability and difficulty in preserving the strict geometric coherence of an aircraft, a limitation that has been extensively documented in surveys on data augmentation for deep
learning~\cite{shorten2019survey} and in studies on the background bias that arises when synthetic samples are not sufficiently diverse~\cite{beery2018recognition}. The recent breakthrough in this area is the use of \glspl{LDM}~\cite{rombach2022high}, which produce images by iteratively denoising a latent representation and achieve
clearly increased photometric fidelity to their predecessors. Adapting a diffusion model with billions of parameters to a specific aircraft class would normally require supercomputing resources, but the \gls{LoRA} technique~\cite{hu2021loralowrankadaptationlarge} makes this feasible by freezing the pretrained weights and injecting
low-rank trainable matrices, which reduces the trainable parameter count by more than 99\,\%. The same generative principle has recently been applied to minority-class augmentation: Mueller et al.~\cite{mueller2024diffusion} proposed an attention-enhanced conditional-diffusion model for data synthesis in machine fault diagnosis, reporting consistent gains for the under-represented classes. The workflow proposed in the present paper transports this idea to the aerospace domain, fine-tuning a state-of-the-art diffusion model with a custom \gls{LoRA} token on the minority helicopter class and pseudo-labelling the synthetic output with an intermediate detector, an integration that, to the authors' knowledge, has not been previously reported.

\subsection{Object detection in aerial imagery}\label{sec:rw_detection}

Before the popularisation of neural networks, aircraft detection in aerial imagery relied on hand-crafted features, exemplified by the combination of \gls{HOG} descriptors~\cite{dalal2005histograms} with \gls{SVM} classifiers~\cite{cortes1995support}. These early models sought geometric silhouettes, projected shadows or sharp edges of aircraft on runways, but they were extremely rigid: they failed under illumination changes, partial occlusion by clouds and variations in the acquisition angle, and produced high false-positive rates against infrastructure of similar shape. The introduction of \glspl{CNN} allowed the network itself to learn geometric and textural features during training, marking the transition towards two-stage detectors such as \gls{R-CNN} and, later, Faster \gls{R-CNN}~\cite{ren2016fasterrcnnrealtimeobject}, in which a \gls{RPN} first proposes candidate regions and a classifier then
confirms the presence of the target. The publication of the \gls{DOTA} dataset~\cite{xia2018dota} consolidated this line by providing hundreds of thousands of annotated aerial instances with Faster \gls{R-CNN} as a baseline. Although two-stage detectors achieved unprecedented precision, their bifurcated architecture was computationally expensive: while one-stage detectors such as \gls{YOLO}~\cite{redmon2016you} reached more than 30 \gls{FPS}, Faster \gls{R-CNN} typically remained below 10 \gls{FPS}, hindering its use in embedded real-time applications. To mitigate the extreme imbalance between background and small targets that characterises aerial scenes, Lin et al.~\cite{lin2017focal} introduced the Focal Loss in RetinaNet, a contribution that directly motivated the present work, since the \textit{small object detection} problem is central to aircraft recognition from \gls{LEO}. Modern \gls{YOLO} versions have since become the industrial standard for aerial imagery, and a number of recent studies in this very journal have explored their behaviour for aircraft detection: Liu et al.~\cite{liu2020aircraft} proposed a corner-clustering approach coupled with deep learning for aircraft detection in remote-sensing imagery, \.Ilmak et al.~\cite{ilmak2025yolo} assessed \gls{YOLO}v8 and v9 for efficient plane detection in very high resolution imagery, and Liu et al.~\cite{liu2026yolo_review} provided a comprehensive review of \gls{YOLO}-based detection in remote sensing. Regarding the small-object regime specifically, Zhang et al.~\cite{zhang2024small} introduced an edge-aware neural network that  explicitly addresses the low signal-to-background ratio of aerial targets. The present work builds on this trajectory but constrains the design from the outset by the \gls{SWAP} budget of a nanosatellite, a condition absent from the aforementioned studies.

\subsection{Synthesis and research gap}\label{sec:rw_gap}

In summary, the reviewed literature shows that one-stage detectors, particularly modern \gls{YOLO} variants, have become the de facto standard for aircraft detection in aerial imagery, that generative data augmentation based on diffusion models and low-rank adaptation has emerged as a viable alternative to classical augmentation for minority classes, and that on-board edge accelerators have reached a maturity sufficient for satellite deployment. However, these three threads have evolved in isolation: no previous work integrates \gls{SWAP}-aware architecture selection, generative data augmentation of the minority class and on-board \gls{INT8} deployment into a single workflow for airborne surveillance from nanosatellites. The present paper addresses precisely that gap and validates the resulting pipeline on a public aerial dataset with full source-code reproducibility.
\section{Mission and Platform Constraints}\label{sec:platform}

This section sets out the physical and operational context that
constrains the rest of the workflow. The nanosatellite platform is
presented first, together with the data-volume bottleneck that
motivates on-board processing; the on-board \gls{AI} accelerator and
its memory constraint are then characterised; and finally the
communications subsystem and the concept of operations that close the
loop with the ground segment are described. The architecture selection
in Section~\ref{sec:methods} and the deployment analysis in
Section~\ref{sec:results} are both driven by the constraints collected
here.

\subsection{CubeSat form factor and orbital window}\label{sec:platform_formfactor}

Historically, space observation has been dominated by monolithic
satellites with development cycles often exceeding a decade,
multi-million-dollar budgets and operational masses of several
tonnes~\cite{wertz2011space}. Over the last two decades, however, the
constant miniaturisation of electronics, the adoption of
\gls{COTS} components and the progressive cost reduction of launch
systems have driven a genuine revolution in space
architecture~\cite{sweeting2018modern}. Within this transition, the
\textit{New Space} paradigm has categorised small satellites by mass,
and nanosatellites platforms between 1 and 10~kg now occupy a
fundamental niche for emerging orbital services~\cite{yost2024state}.
The \gls{CubeSat} standard has been central to this evolution: its
predictable geometric units simplify the design of internal subsystems
and enable deployment through standardised dispensers such as the
P-POD~\cite{puig2001development}, which allows nanosatellites to ride
to orbit as secondary payloads and reduces launch costs by several
orders of magnitude~\cite{katuntsev2010approaches}.

In this work, a 6U \gls{CubeSat}, approximately $10\times20\times30$~cm,
$\sim$8~kg, Table~\ref{tab:cubesat_specs}, is adopted as the reference
platform. Although the \gls{AI} accelerator itself has a minimal
footprint, the 6U format is justified by the optical payload: training
the reference detector at a native resolution of
$1024\times1024$~pixels demands high-resolution imagery, and the 6U
chassis provides the volume required to host the optics and a
sufficiently large solar-panel area for the power budget. The platform
is assumed to operate in a \gls{LEO} at roughly 500~km altitude, where
the satellite travels at approximately 7.5~km/s and the visibility
window over a given ground station is limited to 10--15~minutes per
pass~\cite{ma2020asymnvm}.

\begin{table}[htbp]
\centering
\caption{Dimensional and mass specifications of the \gls{CubeSat}
standard.}
\label{tab:cubesat_specs}
\begin{tabular}{@{}lcc@{}}
\toprule
\textbf{Format} & \textbf{Dimensions (cm)} & \textbf{Max. mass (kg)} \\
\midrule
1U & $10\times10\times10$ & $\sim$1.33 \\
2U & $10\times10\times20$ & $\sim$2.66 \\
3U & $10\times10\times30$ & $\sim$4.00 \\
6U & $10\times20\times30$ & $\sim$8.00 \\
\bottomrule
\end{tabular}
\end{table}

\subsection{Data-volume bottleneck and the case for on-board
processing}\label{sec:platform_bottleneck}

The operational success of \gls{LEO} constellations has triggered a new
transmission challenge. Miniaturised high-resolution optical and
infrared sensors generate an unprecedented volume of information, and
the capacity to acquire photographic telemetry on orbit now exceeds by
far the capacity of the downlink to deliver it to the
ground~\cite{ma2020asymnvm}. This is not an arbitrary design problem
but a barrier imposed by electromagnetics and by the \gls{SWAP} budget
of the platform: a 3U--6U \gls{CubeSat} typically generates only
10--30~W through its solar panels~\cite{poghosyan2017cubesat}, which
restricts the transmit power of the radio-frequency amplifier, and the
reduced volume prevents the installation of high-gain parabolic
antennas. As a result, the mission must rely on low-profile patch
antennas~\cite{gao2014circularly, costantine2016uhf}, whose throughput although sufficient to transmit metadata in real time --- remains
inadequate to sustain a continuous dump of raw high-resolution frames.

This physical limitation consolidates the central proposal of this
work: by integrating a one-stage detector on the on-board processor,
the downlink ceases to be a funnel saturated by gigabytes of pixels
and is used efficiently to transmit only lightweight metadata vectors target coordinates, confidence level and timestamp --- of a few
kilobytes. The contrast with the conventional \textit{bent-pipe}
architecture, in which the satellite acts as a passive collector, is
summarised in Table~\ref{tab:edge_vs_bentpipe}, on-board inference
turns a decision latency of hours or days into fractions of a second
and frees the downlink for tactically relevant information only.

\begin{table*}[htbp]
\centering
\caption{Operational comparison between the conventional bent-pipe
architecture and the on-board edge processing proposed in this work.}
\label{tab:edge_vs_bentpipe}
\begin{adjustbox}{max width=\textwidth}
\begin{tabular}{@{}p{3.2cm} p{5.5cm} p{5.5cm}@{}}
\toprule
\textbf{Operational parameter} & \textbf{Conventional (bent pipe)} &
\textbf{Proposed (edge)} \\
\midrule
On-board processing & None (passive capture and local storage) &
Advanced (active inference with deep learning) \\ \addlinespace
Transmitted data & Raw full images (gigabytes) & Metadata vectors and
detection alerts (kilobytes) \\ \addlinespace
Bandwidth usage & Total saturation of the downlink. & Minimal, avoiding
transceiver bottlenecks \\ \addlinespace
Decision latency & Hours or days (ground post-processing) &
Fractions of a second (near-real-time on orbit) \\ \addlinespace
Energy impact & Sustained high consumption of the RF amplifier &
Optimised, the \gls{TPU} cost is offset by massive RF savings \\
\bottomrule
\end{tabular}
\end{adjustbox}
\end{table*}

\subsection{On-board AI accelerator and memory constraint}\label{sec:platform_tpu}

Moving inference on board requires an accelerator compatible with the (SWaP) budget of a CubeSat, while remaining 
resilient to the radiation environment of Low Earth Orbit (LEO). 
Data-centre Graphics Processing Units (GPUs), such as the NVIDIA T4 
used for the training experiments in this work, consume a peak power 
of 70~W~\cite{nvidia2018t4}. This thermal and electrical footprint is 
fundamentally incompatible with the total 10--30~W power budget 
typically available on a standard nanosatellite platform~\cite{poghosyan2017cubesat}. 
The Commercial Off-The-Shelf (COTS) micro-accelerator landscape evaluated 
for space-borne neural inference comprises three main candidates, 
summarised in Table~\ref{tab:hw_edge}, the Intel Movidius Vision 
Processing Unit (VPU) (Myriad 2/X), the Google Coral Edge Tensor 
Processing Unit (TPU), and the NVIDIA Jetson family (Nano/TX2i/Orin). 
The first two have already been qualified or are under qualification 
by the European Space Agency (ESA): the Myriad 2 was demonstrated on 
orbit by the $\varphi$-sat-1 mission for cloud screening~\cite{rapuano2021fpga}, 
and the Coral Edge TPU is currently being assessed for radiation tolerance 
(Total Ionizing Dose and Single-Event Effects) by the ESA CAIRS21
project~\cite{lentaris2023performance, chanoui2025trends}.

\begin{table*}[htbp]
\centering
\caption{Technical comparison of the COTS micro-accelerators
evaluated for on-board inference.}
\label{tab:hw_edge}
\begin{adjustbox}{max width=\textwidth}
\begin{tabular}{@{}p{3.5cm} p{2.5cm} p{2.0cm} p{2.5cm} p{3.5cm}@{}}
\toprule
\textbf{Platform / chip} & \textbf{Architecture} & \textbf{Power} &
\textbf{Optimal precision} & \textbf{Space validation} \\
\midrule
Intel Movidius (Myriad 2/X) & VPU (vision) & $\sim$1.5--2~W &
FP16 / INT8 & $\varphi$-sat-1 (ESA, 2020) \\ \addlinespace
Google Coral (Edge TPU) & ASIC (tensor) & $\sim$2~W &
INT8 (quantised) & CAIRS21 (ESA, 2024) \\ \addlinespace
NVIDIA Jetson (Nano/TX2i) & SoC (GPU, CUDA) & 5--15~W &
FP32 / FP16 & Aitech Venus space computers \\
\bottomrule
\end{tabular}
\end{adjustbox}
\end{table*}

The Coral Edge TPU is selected as the reference accelerator for
the remainder of this work. It delivers up to 4 Tera-Operations Per 
Second (TOPS) at 2~W (2~TOPS/W)~\cite{prokscha2023efficient}, and its 
compilation ecosystem provides a robust cross-compilation flow that 
maps a model trained on a general-purpose GPU to the on-board
Application-Specific Integrated Circuit (ASIC)~\cite{coral_edgetpu_models_intro}: 
the network is first exported and quantised to INT8 into a \texttt{.tflite}
file~\cite{ultralytics_yolo_export_2024, tensorflow_post_training_integer_quant}, 
and the Edge TPU Compiler then assigns the matrix operations directly 
to the chip instructions. Two strict physical constraints imposed by 
this device drive the architecture selection in Section~\ref{sec:results}:
\begin{itemize}
  \item \textbf{INT8 quantisation.} The accelerator lacks the silicon
  to process 32-bit floating-point (FP32) arithmetic; the entire model 
  must be quantised to 8-bit integers. This step reduces the parametric 
  weight by approximately 75\,\% with only a marginal impact on overall 
  detection precision~\cite{tensorflow_post_training_integer_quant}.
  \item \textbf{8~MB on-die SRAM limit.} The Edge TPU integrates a
  fast but physically constrained Static Random-Access Memory (SRAM) 
  of 8~MB directly on the same die as the processor~\cite{cass2019taking}. 
  For optimal performance, the quantised model's parameters and execution 
  graph must fit entirely within this budget, allowing the full inference 
  path to run in-cache at maximum speed. If the model exceeds this limit, 
  blocks of computation must be offloaded to the general-purpose host 
  CPU of the CubeSat. This offloading triggers massive data transfers 
  across the system bus, causing a severe drop in Frames Per Second (FPS) 
  and negating the energy benefits of the accelerator~\cite{coral_edgetpu_models_intro}.
\end{itemize}

The combination of these two constraints defines the central design
rule of this paper, the selected detector must, once quantised to
INT8, fit strictly within the 8~MB SRAM of the Edge TPU to guarantee 
real-time, energy-efficient performance on orbit.

\subsection{Optical payload and attitude control}\label{sec:platform_optics}

Image acquisition is delegated to \gls{COTS} optics specific to the
\textit{New Space} sector. Modular optical systems such as the
\textit{xScape200} series by Simera Sense~\cite{simera_multiscape200}
or the \textit{Mantis} sensor by Dragonfly
Aerospace~\cite{dragonfly_mantis} integrate organically into a 6U
volume and provide a \gls{GSD} of approximately 1.5~m/pixel from a
500~km orbit. At this resolution, a standard-sized aircraft occupies a
sufficient pixel area on the focal plane to preserve its morphology,
which justifies the native input resolution adopted for the detector. To guarantee orthogonal framing of the ground targets, the satellite
relies on a three-axis \gls{ADCS} combining Star Trackers and Reaction
Wheels, maintaining a strict nadir-pointing attitude during the
acquisition phase and eliminating geometric distortions due to oblique
perspective~\cite{tensortech_adcs_over30kg}.

\subsection{Communications and concept of operations}\label{sec:platform_conops}

\begin{figure}[htbp]
    \centering
    \includegraphics[width=0.95\columnwidth]{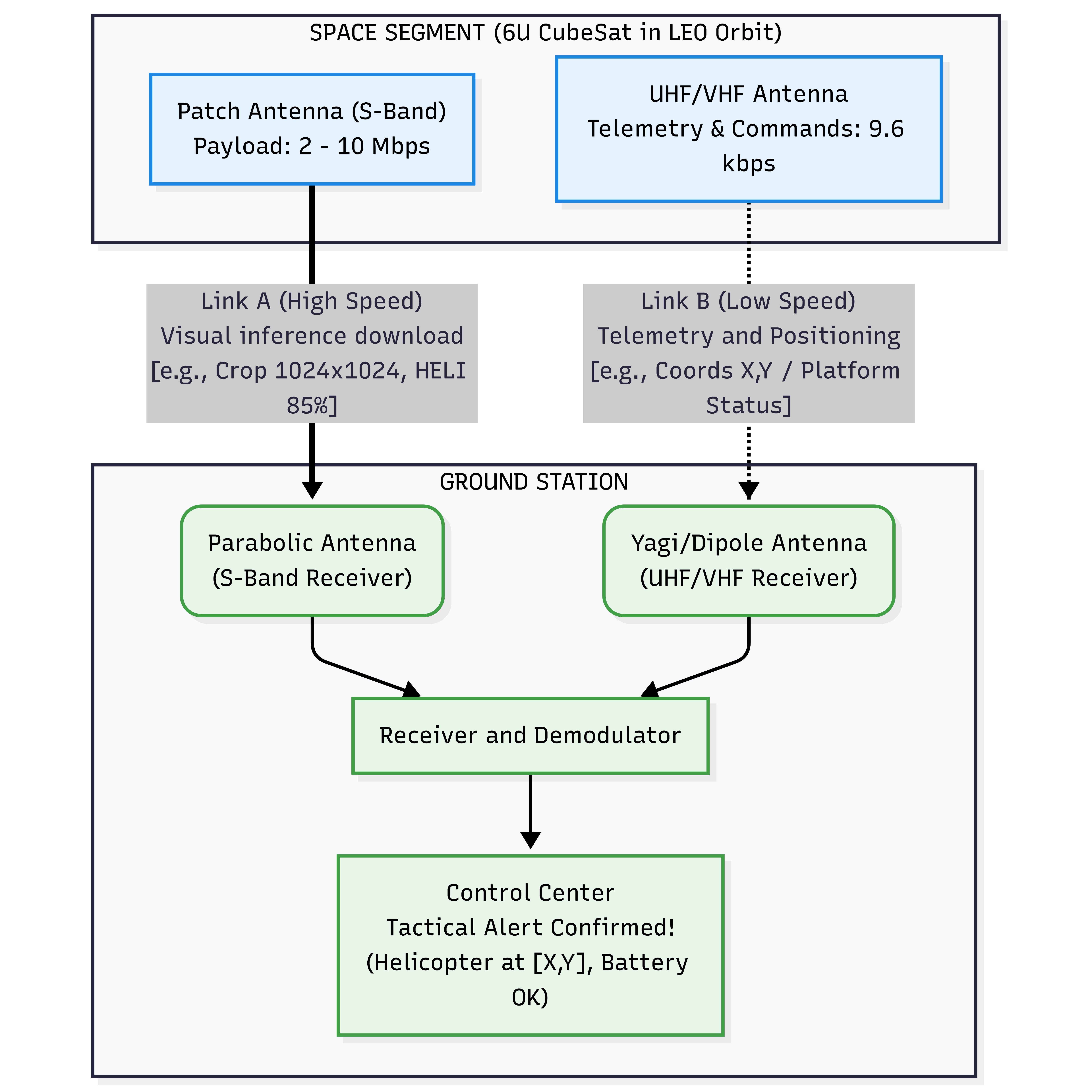}
    \caption{Block diagram of the space-to-ground data-downlink architecture for the 6U CubeSat. The schematic illustrates the dual-band communication strategy: a high-speed S-band link (2--10~Mbps) for visual inference downloads and a low-speed UHF/VHF link (9.6~kbps) for platform telemetry and tactical alerts.}
    \label{fig:downlink_architecture}
\end{figure}

Because the on-board neural processor autonomously discards empty
captures, the volume of data to be transmitted is drastically reduced,
the \gls{CubeSat} only downlinks a lightweight metadata packet, coordinates, timestamp and confidence, together with the compressed
$1024\times1024$-pixel patch corresponding to a positive detection.
A split communications architecture is therefore proposed, illustrated
in Figure~\ref{fig:downlink_architecture}.
\begin{itemize}
  \item \textbf{Payload downlink.} A \gls{COTS} S-band transmitter
  (e.g.\ EnduroSat S-Band Transmitter~\cite{electronics12204374})
  operating in the 2.2--2.3~GHz band provides up to 10~Mbps, which is
  more than sufficient to deliver the detection outputs during the few
  minutes of visibility over the ground station. A low-profile patch
  antenna is used so that the radiating element does not interfere with
  the solar-panel area.
  \item \textbf{\gls{TT&C}.} A standard omnidirectional UHF transceiver
  is retained for command uplink and health-monitoring
  telemetry~\cite{ericsson_digital_airspace}.
\end{itemize}

The mission operates under a total-decision autonomy model that
alternates between low-power and tactical-intervention
modes~\cite{kuwahara_cdhs_2021}:
\begin{itemize}
  \item \textbf{Idle and planning.} The satellite orbits with the
  optical and computational payload switched off. It activates only
  when approaching pre-programmed geographic coordinates of strategic
  interest, located through the on-board \gls{GNSS} receiver.
  \item \textbf{Dynamic acquisition with circular buffer.} While
  overflying the target zone, the camera does not store captures
  permanently; the high-resolution data stream is buffered temporarily
  in the \gls{RAM} of the main on-board computer.
  \item \textbf{Intelligent transmission.} The \gls{AI} accelerator
  analyses the volatile buffer in real time. Frames without detections
  above the confidence threshold are discarded and overwritten
  immediately; upon a positive detection, the system interrupts the
  erase cycle, extracts the useful frame, compresses it together with
  its metadata, and stores it permanently for downlink.
\end{itemize}

This methodology ensures that the space segment delivers exclusively
tactical intelligence, maximising the operational autonomy and
avoiding the bandwidth restrictions that would otherwise saturate the
downlink. The constraints collected in this section 6U form factor,
\gls{INT8} quantisation and the 8~MB \gls{SRAM} budget of the Edge
\gls{TPU}, nadir-pointing optics at $\sim$1.5~m \gls{GSD}, and the
S-band/UHF communications split --- are the inputs that drive the
methodology and the experimental campaign presented next.
\section{Materials and Methods}\label{sec:methods}

This section describes the materials and methods that support the
experimental campaign. The baseline dataset and its initial diagnosis
are presented first (Section \ref{sec:methods_dataset}); the two-stage data-augmentation pipeline ---
classical transformations followed by generative synthesis --- is then
detailed (Section \ref{sec:methods_augmentation}); and finally, in Section \ref{sec:methods_arch}, the candidate detection architectures and the
evaluation metrics are introduced, together with the \gls{SWAP}-aware
selection criterion that links methodology to the platform constraints
of Section~\ref{sec:platform}.

\subsection{Dataset: HRPlanesV2}\label{sec:methods_dataset}

The starting point of the study is the public \textit{HRPlanesV2} dataset, hosted on the Roboflow computer-vision platform~\cite{hrplanesv2_dataset}. It provides a base of high-resolution satellite images with structured annotations of aircraft in three operational categories: civil aircraft, military aircraft, and helicopter. The dataset is partitioned into training, validation, and testing subsets, comprising 79\,\%, 13\,\%, and 8\,\% of the total images, respectively. Before any training, the original training partition was diagnosed along two complementary axes that condition the rest of the workflow.

\begin{figure}[htbp]
    \centering
    \includegraphics[width=0.85\columnwidth]{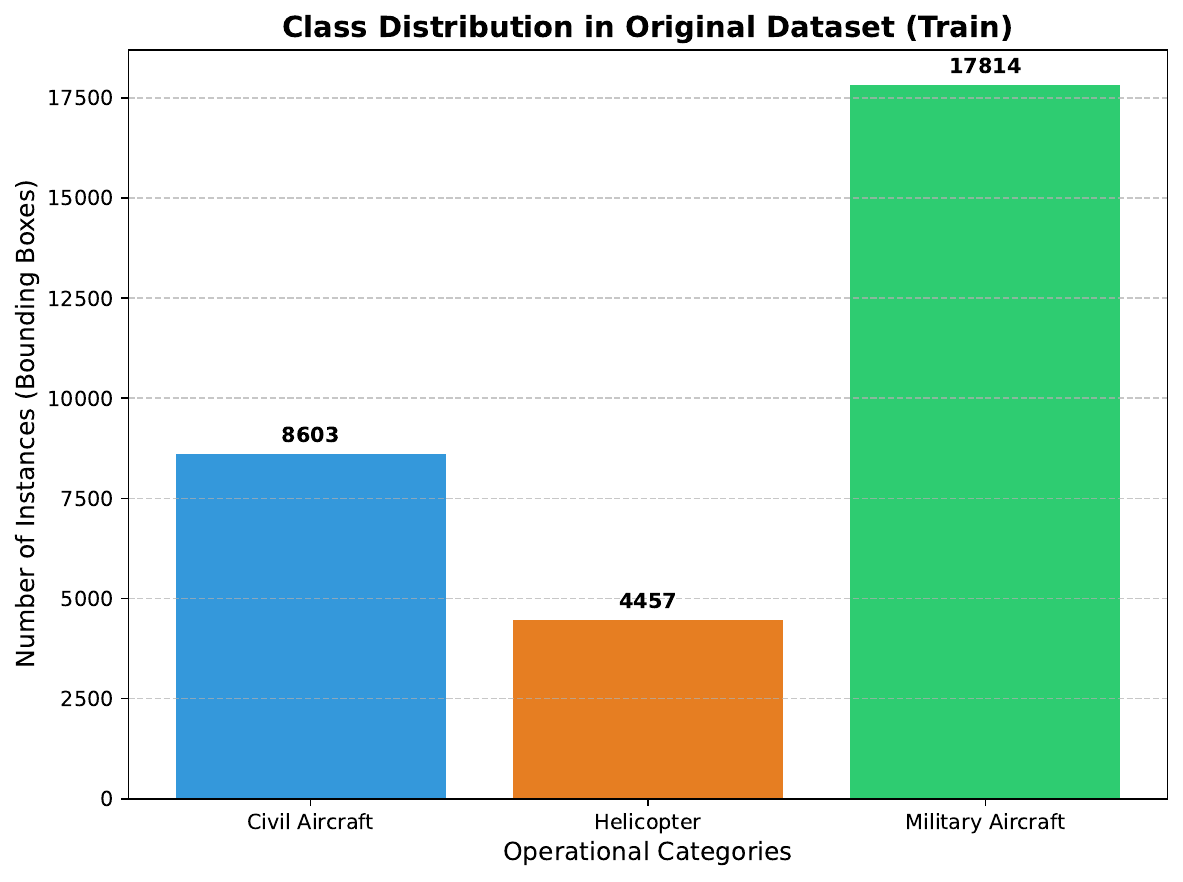}
    \caption{Distribution of instances (bounding boxes) per category
    in the original training set of \textit{HRPlanesV2}.}
    \label{fig:class_distribution}
\end{figure}

The first axis is the quantitative distribution of instances. A total of 30874 valid bounding boxes are available, distributed as follows (see Figure~\ref{fig:class_distribution}): military aircraft is
the majority class with 17814 instances (57.7\,\%), civil aircraft accounts for 8603 instances (27.9\,\%), and helicopters provide only 4457 instances (14.4\,\%). The ratio between the majority and the minority class is approximately 4:1. Training a one-stage detector directly on this distribution biases the network towards the majority class when minimising the global loss,
which is unacceptable for an early-warning mission and motivates the oversampling strategy described next.

\begin{figure}[htbp]
    \centering
    \includegraphics[width=0.85\columnwidth]{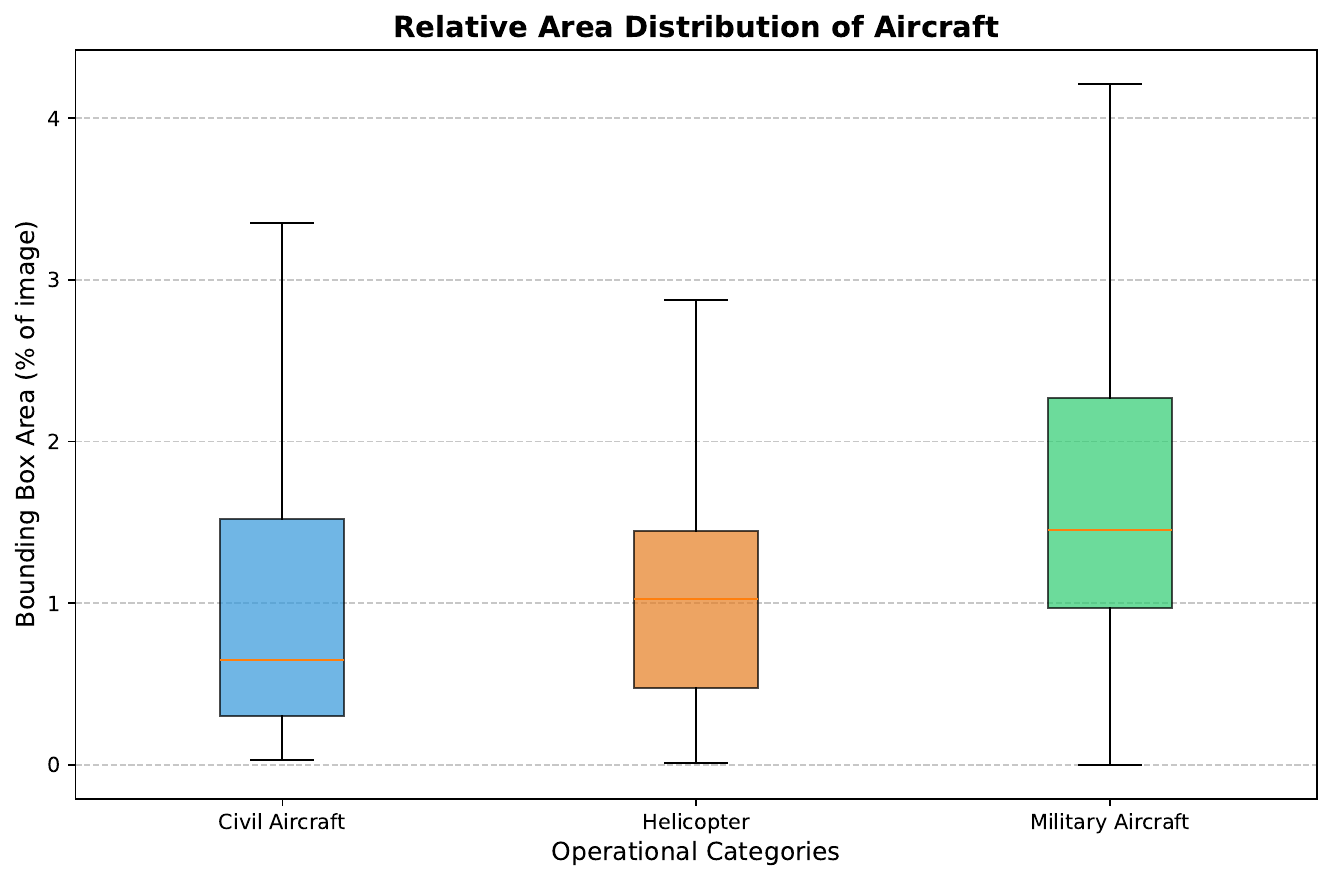}
    \caption{Distribution of the relative area of the bounding boxes
    with respect to the total image size, per operational category.
    Outliers are omitted for readability.}
    \label{fig:area_distribution}
\end{figure}

The second axis is morphological: the relative area of each bounding
box with respect to the full image. A custom Python script was written
to parse every annotation file, extract the bounding-box coordinates
and normalise the geometric area by the image resolution. The
resulting box plot is shown in Figure~\ref{fig:area_distribution}. The
analysis illustrates a classical \textit{small object detection} problem:
62.2\,\% of civil aircraft, 48.4\,\% of helicopters and 26.8\,\% of
military aircraft occupy less than 1\,\% of the total image area. The
helicopter class is the most vulnerable, combining the smallest number
of samples with the smallest mean area (1.08\,\%), against 1.893\,\%
for military aircraft. This indicates that simple whole-image
oversampling is insufficient and that scale-altering transformations
(zoom-in, centred crops) must be included in the augmentation pipeline
to force the network to extract low-level features.

\subsection{Data-augmentation pipeline}\label{sec:methods_augmentation}

To preserve methodological integrity and avoid data leakage, every
transformation described below is applied exclusively to the training
partition; the validation and test partitions are left untouched so
that the final metrics reflect performance on unaltered real scenarios.

\subsubsection{Classical augmentation}\label{sec:methods_classical}

The first stage applies classical transformations to the original
pixels to systematically increase morphological and radiometric diversity 
without requiring additional data collection. Three distinct transformation families 
are combined, all implemented through the widely adopted Albumentations 
library~\cite{buslaev2020albumentations}:
\begin{itemize}
  \item \textbf{Geometric transformations} alter the spatial
  arrangement of pixels without modifying their intensity. Horizontal
  and vertical flips, random rotations and centred crops/zoom-in
  operations are applied. The latter directly addresses the
  small-object problem identified above by forcing the target to
  occupy a larger fraction of the image.
  \item \textbf{Photometric transformations} modify the intensity
  values and colour channels: HSV/RGB jittering, Gaussian and
  salt-and-pepper noise, and Gaussian blur. These emulate the
  radiometric variability that an on-orbit optical payload faces
  across passes (thermal noise, solar glint, partial cloud cover,
  motion blur at 7.5~km/s).
  \item \textbf{Mixed transformations} apply geometric and photometric
  operations sequentially and probabilistically on the same image,
  generating complex edge-case samples that prevent premature convergence and
  reinforce robust feature extraction in the convolutional layers.
\end{itemize}
This stage produced 2832 augmented images (928 geometric, 939
photometric and 965 mixed), bringing the minority classes towards an
approximate balance of 18\,000 instances per class.

\subsubsection{Generative augmentation with FLUX and LoRA}\label{sec:methods_flux}

Classical augmentation is bounded by the information contained in the
original pixels: it can transform existing data but cannot generate
new knowledge. If no original image depicts a helicopter overflying a
snowfield or a desert, no classical transformation can recreate that
scenario. To overcome this representational limit, a generative stage
based on Latent Diffusion Models (LDMs) is introduced. Generative Adversarial 
Networks (GANs)~\cite{goodfellow2014generativeadversarialnetworks}
were discarded because of their training instability and difficulty
in preserving the strict geometric coherence of an aircraft, as
documented in surveys on augmentation for deep
learning~\cite{shorten2019survey} and in studies on the background
bias that arises from insufficiently diverse synthetic
samples~\cite{beery2018recognition}. The selected generative backbone
is FLUX~\cite{blackforest2024flux}, which replaces traditional
diffusion with flow-matching and provides the photorealistic fidelity
required for military fuselages and helicopter rotors.

Adapting a billion-parameter diffusion model to a specific aircraft
class would normally demand supercomputing resources. This is made
feasible by Low-Rank Adaptation (LoRA)~\cite{hu2021loralowrankadaptationlarge}, which
freezes the pretrained weight matrix $W_0$ and approximates the
update $\Delta W$ as the product of two low-rank matrices $B$ and $A$:
\begin{equation}
    W_{\text{new}} = W_0 + B\,A,
    \label{eq:lora}
\end{equation}
where, if $W_0\in\mathbb{R}^{d\times k}$, then $B\in\mathbb{R}^{d\times r}$
and $A\in\mathbb{R}^{r\times k}$ with intrinsic rank $r\ll\min(d,k)$.
For a $10\,000\times10\,000$ matrix (100~M parameters), the
decomposition with $r=8$ reduces the trainable count to
$160\,000$ ($-99.8\,\%$), which makes fine-tuning tractable on
consumer hardware.

\begin{figure}[htbp]
    \makebox[\textwidth][c]{\includegraphics[width=1.1\textwidth]{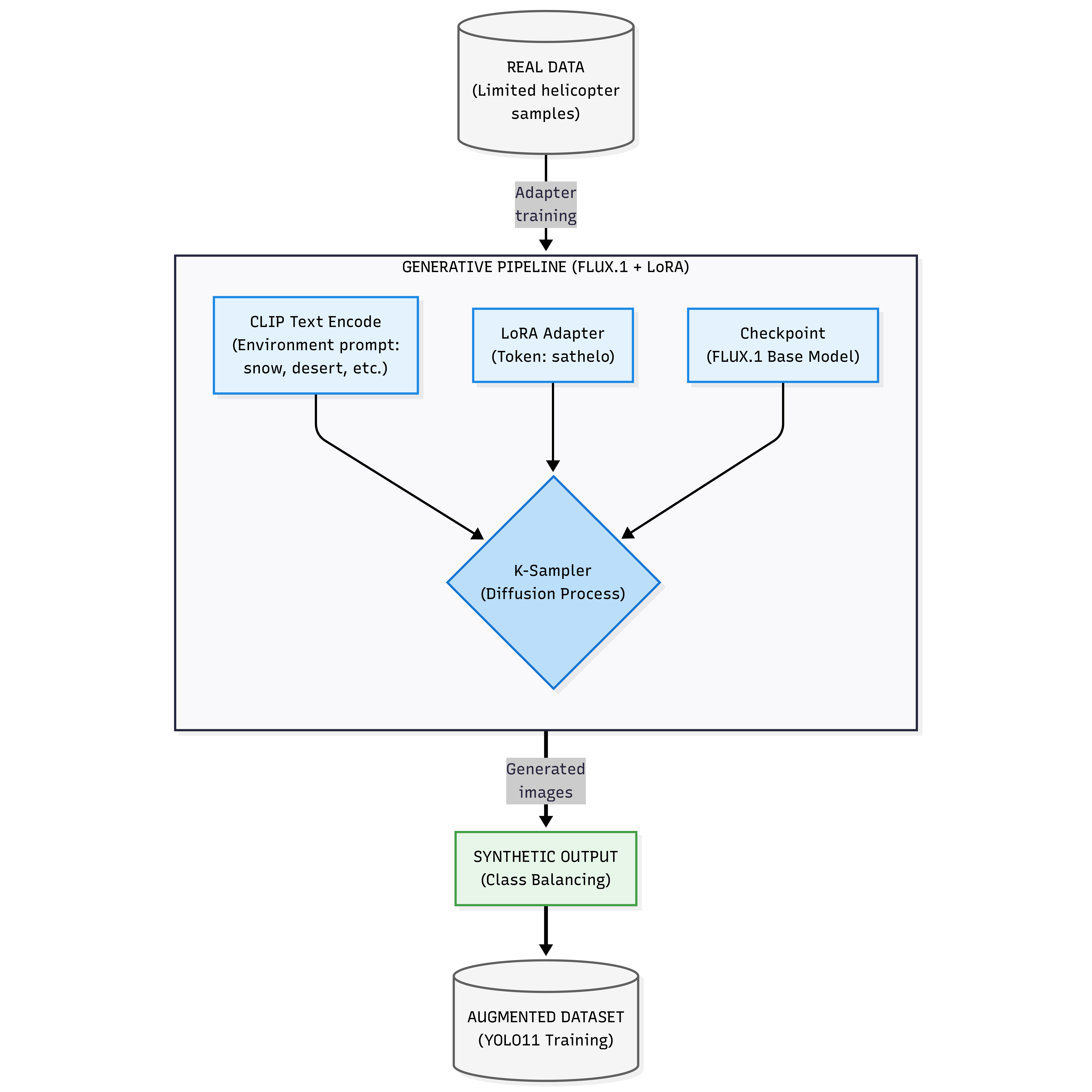}}
    \caption{Block diagram of the generative data-augmentation pipeline. The limited real helicopter instances are used to train a custom LoRA adapter. During inference, the FLUX.1 base model, combined with the domain-specific LoRA and environmental text prompts, feeds the K-Sampler to generate a diverse synthetic output, balancing the final YOLO11 dataset.}
    \label{fig:flux_pipeline}
\end{figure}

The concrete pipeline operates as follows (Figure~\ref{fig:flux_pipeline}). The base model
\textit{FLUX.1-dev} is loaded in quantised GGUF format (Q4\_0) so that
the fine-tuning fits comfortably within the 16~GB of VRAM available
on the university workstation. A custom LoRA
adapter is trained on the minority helicopter class with the trigger
token \texttt{sathelo}. Inference is run through a ComfyUI workflow
composed of a UNET GGUV loader, a dual CLIP loader (CLIP-L and
T5-XXL) for text conditioning, the LoRA adapter applied with
model strength 0.85 and clip strength 1.0, a KSampler configured with
27 steps, CFG~=~1, Euler sampler and Beta scheduler at denoise 0.95,
and a VAE decode stage. Synthetic
helicopter imagery is generated across four operational environments
to maximise contextual diversity: industrial/airport, rural/forest,
coastal/maritime and arid/desert, yielding approximately 2\,226
helicopter images (see Figure~\ref{fig:flux_pipeline}). For the civil
class, classical Albumentations transforms (flip, HueSaturationValue,
RandomBrightnessContrast, GaussNoise, Blur, ShiftScaleRotate) are
applied to existing real samples.

\subsubsection{Pseudo-labelling and class balancing}\label{sec:methods_pseudo}

Manually annotating thousands of synthetic images is unfeasible. An
intermediate \gls{YOLO}v11 detector, previously trained on the
classically augmented set, is used to auto-label the synthetic output. The resulting labels are merged with the classically augmented samples, producing a final balanced dataset of more than 53\,000 instances at approximately
18\,000 instances per class.

\subsection{Candidate architectures and evaluation metrics}\label{sec:methods_arch}

Three one-stage detectors are evaluated as candidates for the on-board
mission, chosen to span the precision--efficiency design space:
\begin{itemize}
  \item \textbf{SSD MobileNet V3 Large}~\cite{liu2016ssd}: a
  depthwise-separable backbone that represents the historical standard
  for ultra-lightweight deployment.
  \item \textbf{YOLO11n}~\cite{yolo11_ultralytics, redmon2016you}: the
  nano version of the latest \gls{YOLO} release, built around a
  C3K2 backbone with \gls{SPPF} and a \gls{PANet} neck, plus a
  C2PSA spatial-attention head.
  \item \textbf{RT-DETR-L}~\cite{zhao2024detrs, carion2020end}: a
  hybrid convolutional/Transformer detector that replaces the
  anchor-based head with an end-to-end bipartite-matching decoder,
  eliminating \gls{NMS}.
\end{itemize}

All models are evaluated with the same metric set, which jointly
captures detection quality and deployability:
mAP@50 and mAP@50-95 for detection quality;
Precision, Recall and F1-Score per class
for tactical behaviour; and parameters, weight on
disk, GFLOPs and FPS for computational cost. The
\gls{SWAP}-aware selection criterion is defined explicitly at this
point and applied in Section~\ref{sec:results}: a candidate is
admissible only if, after \gls{INT8} quantisation, its parametric
weight fits within the 8~MB on-die \gls{SRAM} of the Coral Edge
\gls{TPU} (Section~\ref{sec:platform_tpu}); otherwise the inference
path falls back to the general-purpose \gls{CPU} and the on-orbit
\gls{FPS} collapses. This criterion is the link between the
mission-and-platform constraints of Section~\ref{sec:platform} and the
experimental campaign that follows.
\section{Experiments and Results}\label{sec:results}
This section details the empirical evaluation of the proposed methodology. We first outline the hardware and software environments used for training, alongside the specific hyperparameters and configuration choices governing the experimental pipeline (Section~\ref{sec:results_setup}). Subsequently, we present a comparative benchmark of the three candidate one-stage architectures---SSD MobileNet V3, RT-DETR-L, and YOLO11n---evaluated on the original, unbalanced \textit{HRPlanesV2} dataset to justify the selection of YOLO11n based on its precision, inference speed, and edge-compatibility constraints (Section~\ref{sec:results_benchmark}).

\subsection{Experimental Setup and Hyperparameters}\label{sec:results_setup}
To ensure full reproducibility of the computational experiments, the training configurations were standardized across all evaluated models. The experiments were conducted in a Google Colab environment utilizing a dedicated GPU accelerator. For the final YOLO11n training on the augmented dataset, the network was initialized with pre-trained COCO weights (\path{yolo11n.pt}) to accelerate convergence and subsequently fine-tuned. Table~\ref{tab:hyperparameters} summarizes the primary hyperparameters explicitly configured during the training phase, including the exact random seed used to guarantee identical replication of the reported metrics. Unlisted hyperparameters were kept at their default Ultralytics settings.

\begin{table}[htbp]
    \centering
    \caption{Training hyperparameters for the YOLO11n network.}
    \label{tab:hyperparameters}
    \begin{tabular}{@{}ll@{}}
        \toprule
        \textbf{Hyperparameter} & \textbf{Value} \\
        \midrule
        Input Resolution & $640 \times 640$ pixels \\
        Epochs & 20 \\
        Batch Size & Auto (memory-optimized, \texttt{batch = -1}) \\
        Random Seed & 93 \\
        Initial Weights & \path{yolo11n.pt} \\
        \bottomrule
    \end{tabular}
\end{table}

Preliminary training was carried out on a Google Colab \gls{GPU} NVIDIA T4, and the final runs on an NVIDIA RTX~4080 16~GB workstation. To explicitly rule out stochastic optimization bias and characterize the variability of the network, a five-seed study was conducted (seeds 0, 16, 42, 93, and 2026). The model's performance remains highly stable across initializations, yielding a global mean mAP@50 of 0.805 with negligible variance. Because the architecture is not affected by random weight initialization, seed 93---which closely aligns with the average validation losses and precision-recall trade-off---was systematically selected as the representative model for all subsequent inference and quantization experiments. The full stochastic study is released in the public repository for reproducibility. All models were trained with the AdamW optimiser, an automatic batch size (\texttt{AutoBatch = -1}), \gls{AMP} (FP16), and the mosaic-augmentation schedule recommended by Ultralytics for \gls{RT-DETR}. A resolution sweep between 640 and 1024~pixels was performed for \gls{YOLO}11n to quantify the impact of the input size on small-object detection.

\begin{table*}[htbp]
\centering
\caption{Comparative benchmark of the three candidate one-stage
detectors on the unbalanced \textit{HRPlanesV2} dataset. The Edge
\gls{TPU} admits only models whose \gls{INT8}-quantised weight fits
its 8~MB on-die \gls{SRAM}.}
\label{tab:benchmark}
\begin{adjustbox}{max width=\textwidth}
\begin{tabular}{@{}lccccc@{}}
\toprule
\textbf{Model} & \textbf{Resolution} & \textbf{Params} &
\textbf{Weight} & \textbf{mAP@50} & \textbf{FPS} \\
\midrule
SSD MobileNet V3 & 320 & 3.07~M & 11.93~MB & 44.98\,\% & 82.66 \\
RT-DETR-L & 640 & 32.81~M & 188.85~MB & 74.53\,\% & 14.05 \\
YOLO11n & 640 & 2.59~M & 5.22~MB & 75.01\,\% & 48.28 \\
YOLO11n & 1024 & 2.59~M & 5.22~MB
& 77.90\,\% & 34.46 \\
\bottomrule
\end{tabular}
\end{adjustbox}
\end{table*}

As detailed in Table~\ref{tab:benchmark}, the candidate models were evaluated to determine their suitability for on-board edge deployment. SSD MobileNet V3 achieved the highest inference speed (82.66~FPS) but was discarded due to a severe background collapse, yielding an unacceptable mAP@50 of 44.98\,\%. Conversely, while RT-DETR-L demonstrated strong detection capabilities, it was ruled out because its 188.85~MB memory footprint heavily exceeds the strict 8~MB on-die \gls{SRAM} limit of the Edge \gls{TPU}. Ultimately, YOLO11n was selected as the optimal architecture. Both its 640 and 1024-pixel resolution variants comfortably fit within the hardware memory constraints (5.22~MB). The 1024-pixel configuration was chosen as the definitive baseline, as it maximizes the small-object detection performance (77.90\,\% mAP@50) while maintaining a highly capable real-time inference rate of 34.46~FPS.

\subsection{Architecture benchmark (unbalanced dataset)}\label{sec:results_benchmark}

The three candidate architectures were first trained and evaluated on
the unbalanced \textit{HRPlanesV2} partition described in
Section~\ref{sec:methods_dataset}. The resulting figures are
summarised in Table~\ref{tab:benchmark} and comprehensively analyzed 
across three key visual dimensions: the balance between detection 
capacity and inference speed, the 
hardware constraints regarding memory and parametric load, and 
the operational design space.

\begin{figure}[htbp]
    \centering
    \includegraphics[width=\columnwidth]{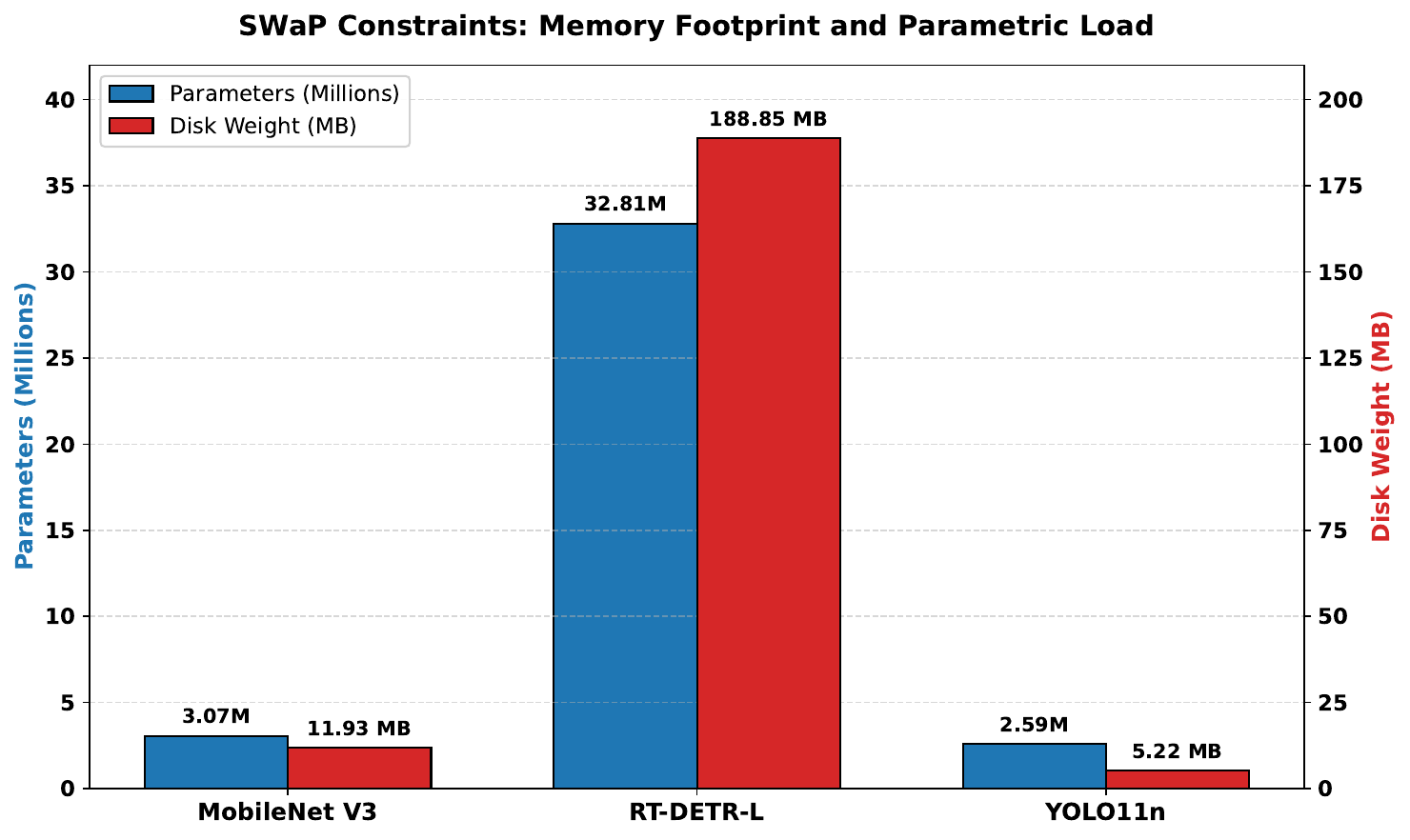}
    \caption{Direct comparison between mean average precision
    (mAP@50) and baseline inference speed.}
    \label{fig:precision_fps}
\end{figure}

\begin{figure}[htbp]
    \centering
    \includegraphics[width=\columnwidth]{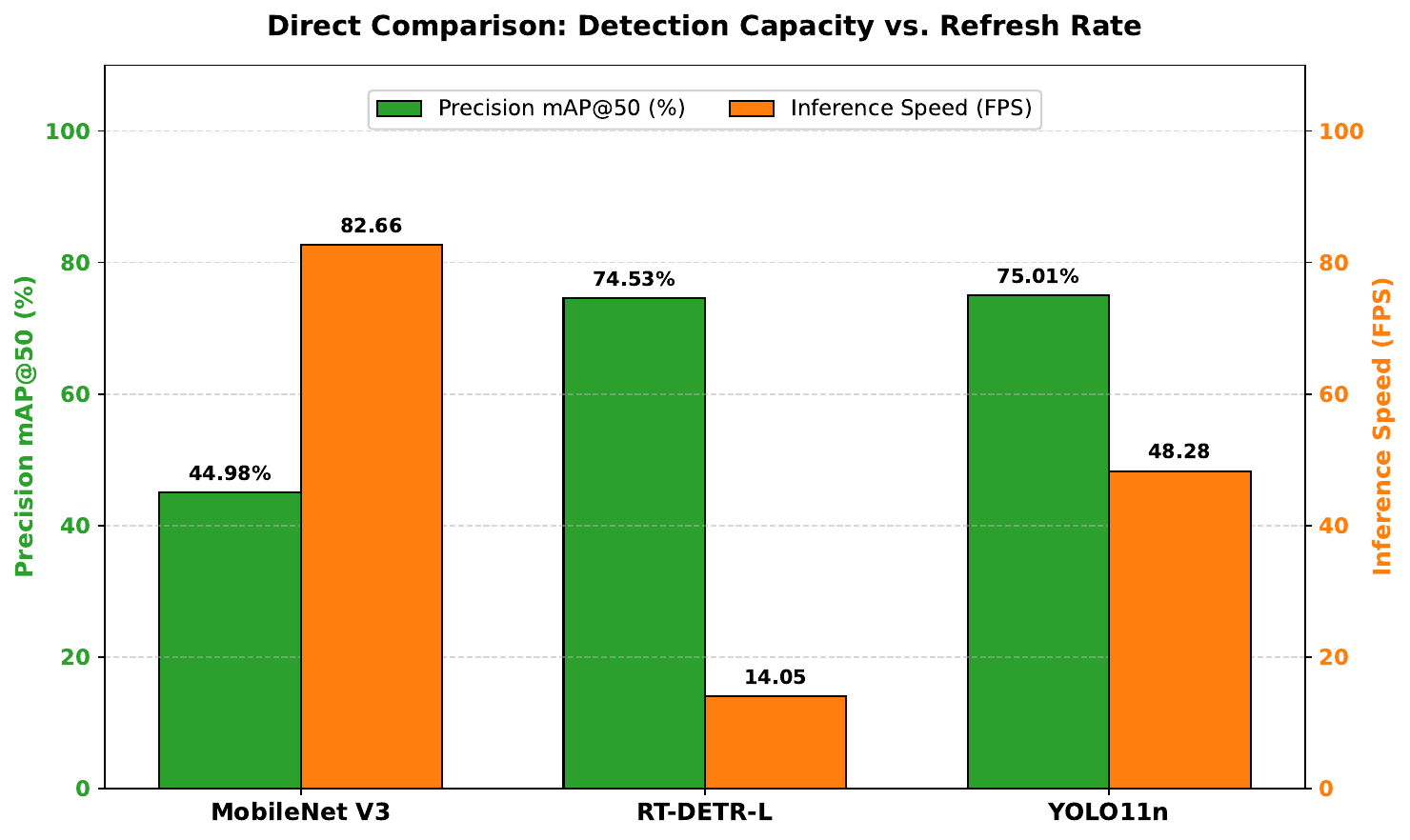}
    \caption{Parametric footprint of each architecture against the
    8~MB \gls{SRAM} budget of the Edge \gls{TPU}. \gls{RT-DETR}-L
    exceeds the limit by nearly 190~MB; \gls{YOLO}11n fits comfortably
    after \gls{INT8} quantisation.}
    \label{fig:memory_footprint}
\end{figure}

As illustrated in Figure~\ref{fig:precision_fps}, SSD MobileNet V3, 
despite being the fastest and lightest candidate (3.07~M
parameters, 11.93~MB), collapses onto the background class with a
mAP@50 of only 44.98\,\%. This critical failure is due to an incompatibility 
between its ImageNet-pretrained backbone and the distinct statistical distribution 
of the aerial scenes. Conversely, \gls{RT-DETR}-L achieves a competitive 
74.53\,\% mAP@50, but as clearly depicted in Figure~\ref{fig:memory_footprint}, 
its 188.85~MB footprint exceeds the 8~MB \gls{SRAM} budget by more
than an order of magnitude. Even after \gls{INT8} quantisation, the
model cannot reside in the accelerator cache, and its
attention-specific matrix operations are not supported by the
integer \gls{ALU} of the Edge \gls{TPU}, forcing a fallback to the
\gls{CPU} and consequently collapsing the inference speed to 14.05~\gls{FPS}. 

\begin{figure}[htbp]
    \centering 
    \includegraphics[width=0.92\columnwidth]{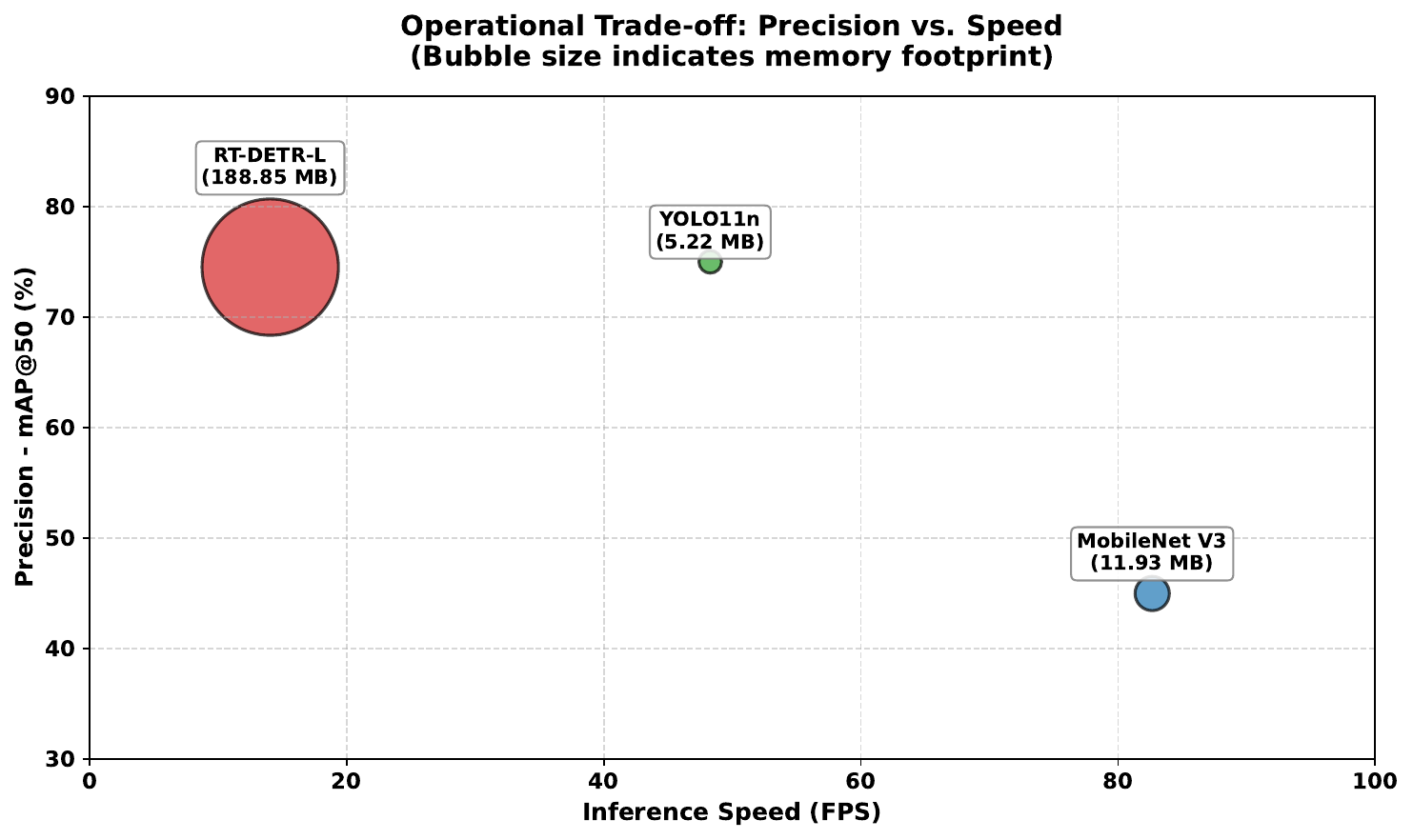}
    \caption{Operational design space. The bubble size is proportional
    to the model weight on disk (MB). \gls{YOLO}11n is the only
    candidate in the upper-right region with a minimum radius.}
    \label{fig:tradeoff_bubbles}
\end{figure}

\gls{YOLO}11n offers the optimal precision--efficiency compromise. 
Figure~\ref{fig:tradeoff_bubbles} visually encapsulates this advantage: 
it is the only candidate that simultaneously reaches the upper-right 
(high precision, high speed) region while maintaining a minimal radius, 
indicative of its negligible memory footprint. At 640~px, it already matches the
precision of the Transformer-based detector at more than three times
the speed (48.28 vs.\ 14.05~\gls{FPS}). Furthermore, increasing the input
resolution to 1024~px lifts mAP@50 to 77.90\,\% and mAP@50-95 by
6.65 percentage points, with a particularly strong impact on the
helicopter class (where the acquisition rate rises from 46\,\% to 57\,\%).
The operating point on the F1 curve is fixed at a confidence
threshold of 0.78, where the F1 score reaches 0.78. 

The residual weakness of the unbalanced \gls{YOLO}11n model lies in the
helicopter class: in the normalised confusion matrix, 34\,\% of
helicopter instances are misclassified as background, and a residual
19\,\% confusion between civil and military aircraft remains. This outlines
the open problem that the augmentation campaign of
Section~\ref{sec:results_augmentation} successfully targets.
\subsection{Impact of data augmentation}\label{sec:results_augmentation}

Two experiments are compared against the unbalanced \gls{YOLO}11n
baseline of Section~\ref{sec:results_benchmark}, both trained for
20 epochs at 1024~px with seed~93.

\subsubsection{Experiment 1: classical augmentation}\label{sec:results_classical}

Applying the classical pipeline of Section~\ref{sec:methods_classical}
yields 2\,832 additional images and brings the classes to an
approximate balance of 18\,000 instances each. Global mAP@50 rises
from 77.90\,\% to 81.62\,\%, and the maximum F1 to 0.77 at a
confidence threshold of 0.264. The per-class F1, however, exposes an
asymmetry: civil aircraft reaches 0.847 and military aircraft 0.887,
but helicopters remain at 0.683, with 31\,\% of helicopter instances
still collapsing onto the background. Classical augmentation is
therefore effective for the majority classes but insufficient for the
minority class.

\begin{figure}[htbp]
    \centering
    \includegraphics[width=\columnwidth]{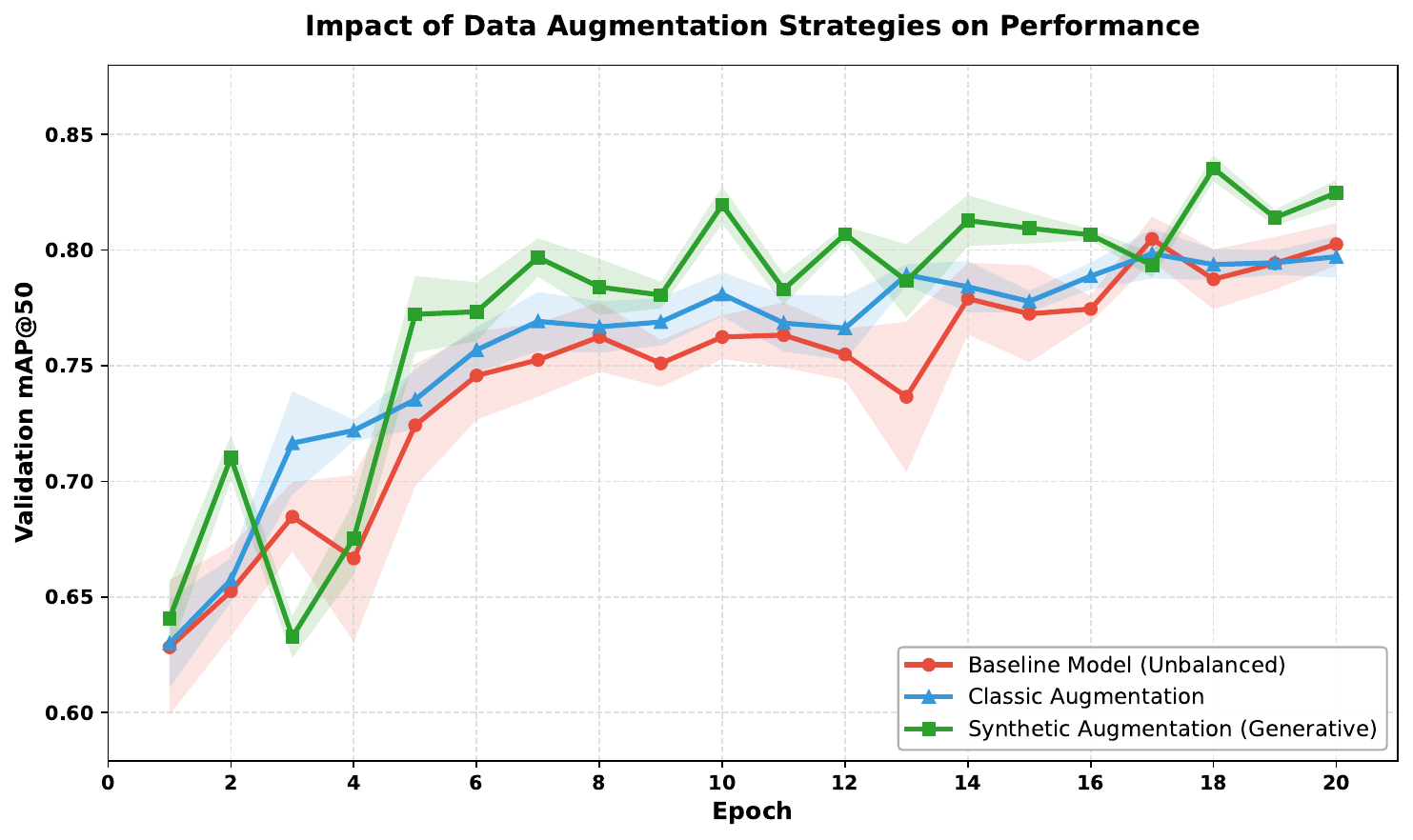}
    \caption{Evolution of mAP@50 across the three training regimes:
    unbalanced baseline, classical augmentation and FLUX+\gls{LoRA}
    generative augmentation.}
    \label{fig:map_balanced}
\end{figure}

\subsubsection{Experiment 2: FLUX + LoRA generative augmentation}\label{sec:results_synth}

\begin{figure}[htbp]
    \centering
    \includegraphics[width=0.82\columnwidth]{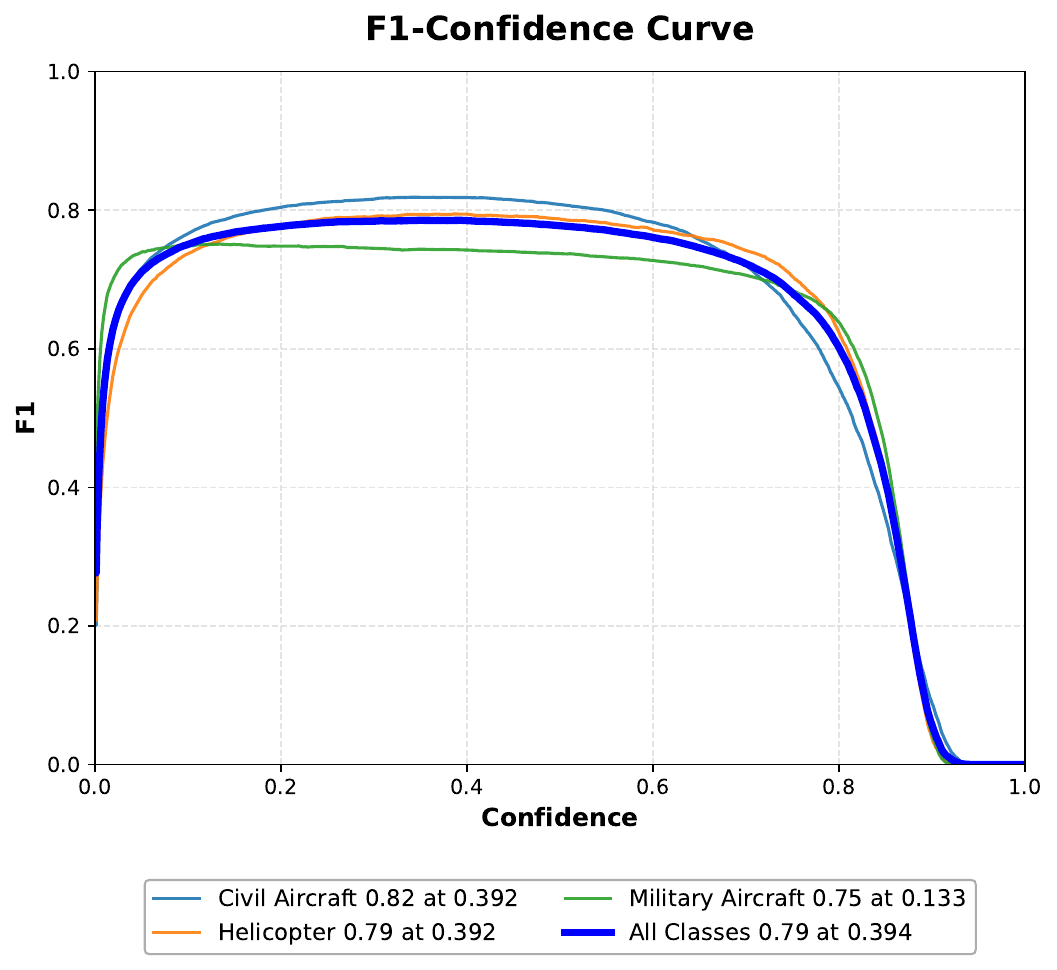}
    \caption{F1 curve of the final \gls{YOLO}11n model trained on the
    FLUX+\gls{LoRA}-balanced dataset, with a maximum F1 of 0.79 at a
    confidence threshold of 0.394.}
    \label{fig:f1_curve}
\end{figure}

The overall performance progression and the optimal operating threshold of the model are further detailed in Figure~\ref{fig:map_balanced} and Figure~\ref{fig:f1_curve}. As illustrated in Figure~\ref{fig:map_balanced}, the evolution of the validation mAP@50 metric across the three training regimes underscores the effectiveness of the proposed methodology. While the transition from the unbalanced baseline to the classical augmentation provided initial improvements, it is the introduction of the FLUX+\gls{LoRA} generative augmentation that maximizes the global performance. Furthermore, the narrow confidence intervals depicted in the learning curves demonstrate that this synthetic balancing is highly robust against stochastic optimization variance. Ultimately, this approach not only increases the global mAP@50 but also ensures a more equitable feature representation across all aeronautical categories, effectively closing the performance gap between the minority and majority classes.

Furthermore, the F1-confidence curve (Figure~\ref{fig:f1_curve}) evaluates the harmonic mean of precision and recall across different confidence thresholds. The model reaches a maximum F1 score of 0.79 at a confidence threshold of 0.394. This specific optimal threshold aligns with the previously discussed synthetic domain shift: the model requires a moderate, rather than extremely high, confidence threshold to maximize the detection of challenging or degraded real-world instances without succumbing to excessive background false positives. Together, these metrics empirically validate the generative augmentation strategy as a highly effective approach for object detection in imbalanced datasets.

\begin{figure}[!h]
    \makebox[\textwidth][c]{\includegraphics[width=0.92\textwidth]{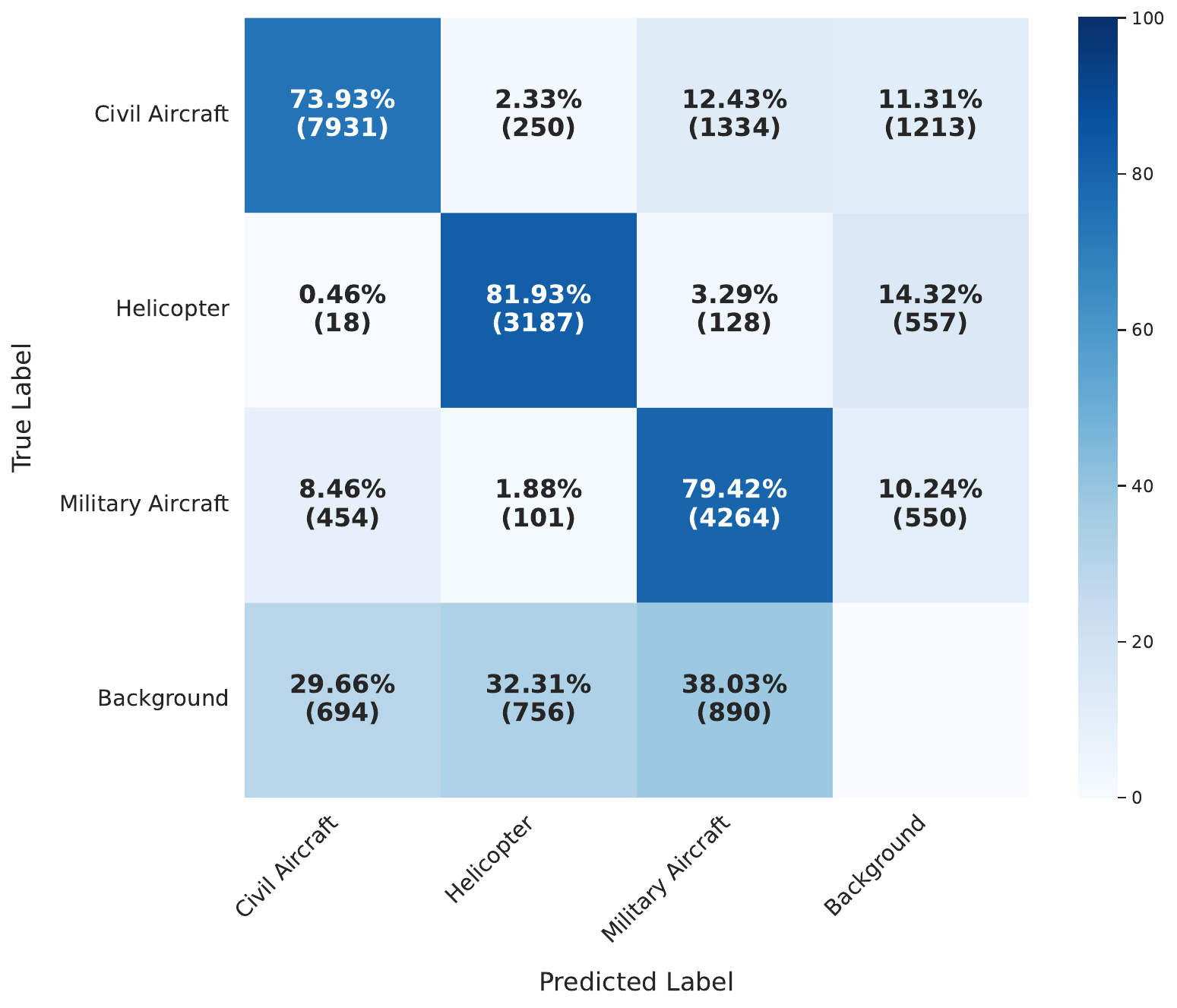}}
    \caption{Combined confusion matrix of the final balanced model, displaying both normalized percentages and absolute instance counts. The diagonal highlights the detection rates across all targeted aeronautical categories.}
    \label{fig:cm_combined}
\end{figure}

To fully characterize the model's behaviour, a detailed analysis of the combined confusion matrix (Figure~\ref{fig:cm_combined}) is required, which presents both normalized percentages and absolute metrics simultaneously. The matrix indicates that the model possesses strong detection capabilities, correctly identifying 7931 civil aircraft (74\%), 3187 helicopters (82\%), and 4264 military aircraft (79\%).

However, particular attention must be given to background misclassifications. The persistence of false negatives---true aeronautical instances predicted as background, totalling 1213 civil, 557 helicopters, and 550 military instances---can be attributed to the nature of the synthetic dataset expansion. While generative augmentation (FLUX+\gls{LoRA}) successfully balances the class distribution and provides high-quality samples, it can introduce a subtle domain shift. The neural network learns highly specific synthetic feature representations that may occasionally fall short of capturing the full complexity, varied lighting, or sensor noise of extreme real-world edge cases. Consequently, when faced with highly degraded real-world instances, the model adopts a conservative inference strategy, abstaining from predictions and defaulting to the background class.

\begin{figure}[htbp]
    \makebox[\textwidth][c]{\includegraphics[width=0.96\textwidth]{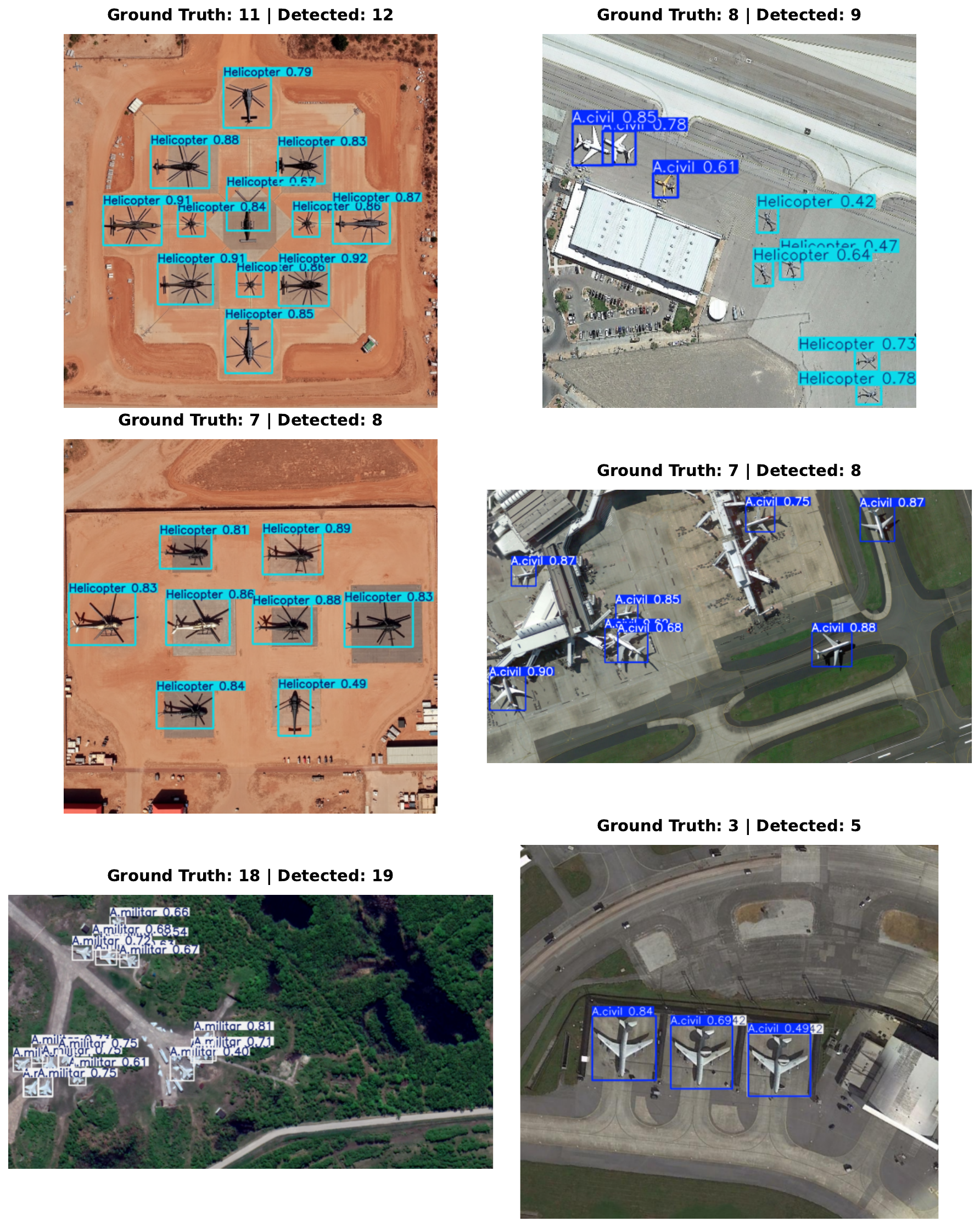}}
    \caption{Qualitative evaluation of apparent false positives. The inference results show scenes where the trained model successfully detects valid aircraft that were entirely missing from the original ground-truth annotations (e.g., predicting 14 objects where the dataset only labeled 13). This supports that a portion of the reported background misclassifications are actual correct detections, highlighting the model's feature extraction capabilities.}
    \label{fig:unannotated_detections}
\end{figure}

Conversely, the false positive distribution (background regions predicted as aircraft, comprising 694 civil, 756 helicopters, and 890 military instances) highlights the model's heightened feature-extraction sensitivity. This phenomenon is largely driven by two factors. First, many of these perceived errors are overlapping bounding boxes on a single aircraft, where the evaluation metric penalizes the duplicate as a background false positive. Second, the dataset contains originally unlabeled instances that human annotators missed. The current detector successfully identifies these unannotated targets, effectively outperforming the original ground-truth annotations in certain scenes. This capability is visually corroborated in subsequent qualitative evaluations (e.g., Figure~\ref{fig:unannotated_detections}), which show the model detecting valid aircraft that were completely absent from the original dataset labels.

Regarding the background class, it is important to note that the intersection representing True Negatives (Background-Background) is left intentionally blank. Unlike standard image classification, object detection frameworks evaluate performance based on discrete bounding box proposals. Consequently, the background comprises a virtually infinite number of spatial locations across the image where no object exists and none is predicted. Since True Negatives are mathematically unquantifiable in this spatial context and do not contribute to standard detection metrics such as Precision, Recall, or mAP, this cell is omitted to prevent the distortion of the overall performance statistics.

\subsection{Deployable model: INT8 quantisation and on-orbit
inference}\label{sec:results_deployment}

Once \gls{YOLO}11n is selected, the model is exported through the
Ultralytics export pipeline~\cite{ultralytics_yolo_export_2024} to a
\texttt{.tflite} file and quantised to \gls{INT8} with the
post-training integer-quantisation toolchain of
TensorFlow~\cite{tensorflow_post_training_integer_quant}. The Edge
\gls{TPU} Compiler then maps the matrix operations directly to the
chip instructions. Quantisation reduces the parametric weight from
5.22~MB to approximately 2.5~MB ($-75\,\%$), so the entire inference
path resides in the 8~MB on-die \gls{SRAM} and runs without
\gls{CPU} offloading. Scaling the measured T4 latency to the Edge
\gls{TPU} throughput-to-power ratio yields a projected on-orbit rate
of 25--30~\gls{FPS} at 1024~px.

To handle full-scene acquisitions without degrading the native \gls{GSD},
inference is wrapped in a \gls{SAHI} scheme~\cite{akyon2022slicing}:
each high-resolution capture is sliced into $1024\times1024$ patches
with a 15\,\% perimeter overlap, the Edge \gls{TPU} evaluates every
patch sequentially, and the validated detection vectors are projected
back onto the global coordinate system of the original capture, where
\gls{NMS} merges duplicate predictions in the overlap regions.

\subsection{Extended multi-class model and ISR application}\label{sec:results_extended}

\begin{figure}[htbp]
    \makebox[\textwidth][c]{\includegraphics[width=1.3\textwidth]{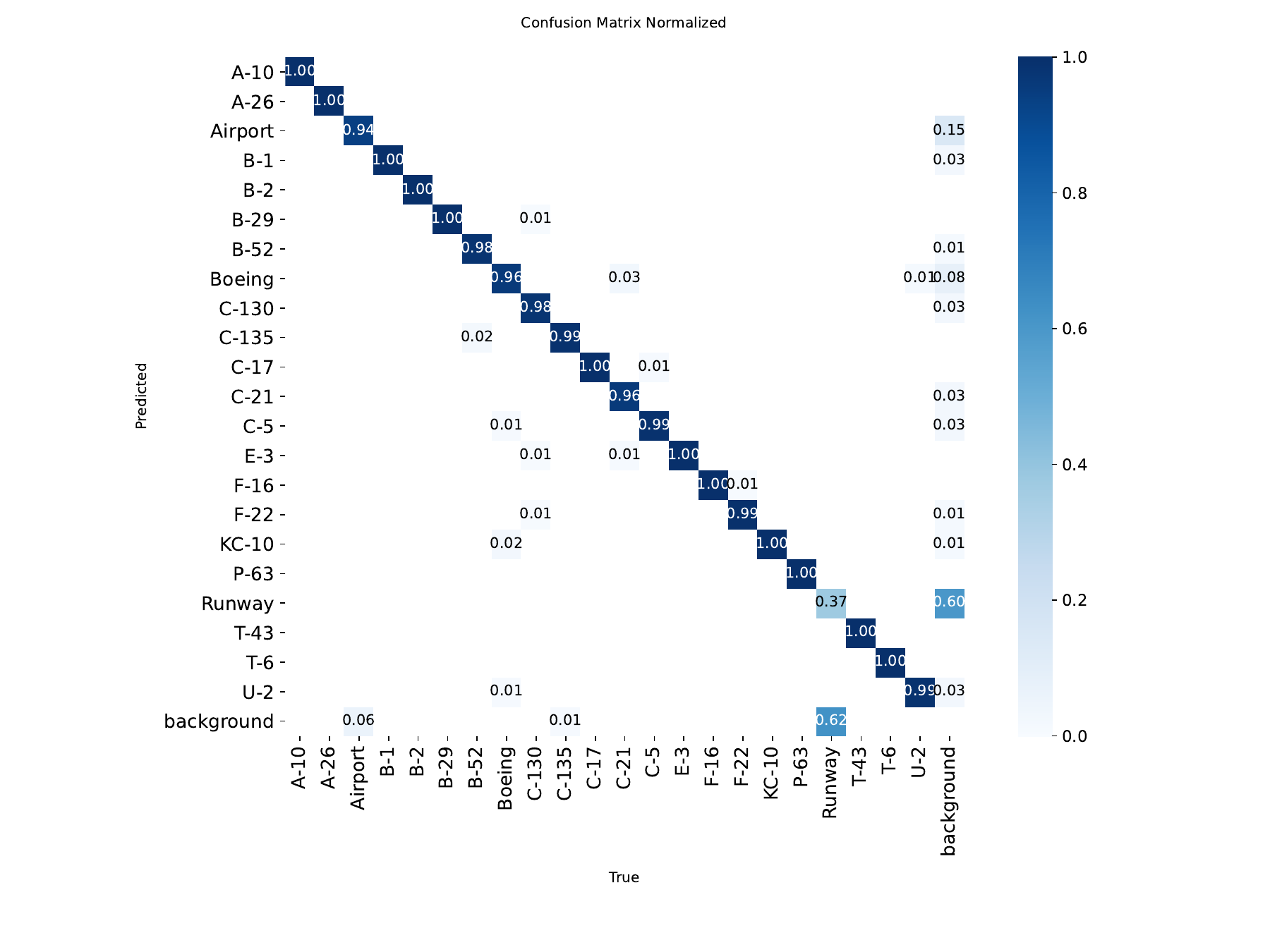}}
    \caption{Normalised confusion matrix of the extended 22-class
    YOLO11n model. The diagonal exceeds 95\,\% for almost every
    class except \textit{Runway} (37\,\%).}
    \label{fig:cm_22class}
\end{figure}

\begin{figure}[htbp]
    \centering
    \includegraphics[width=\columnwidth]{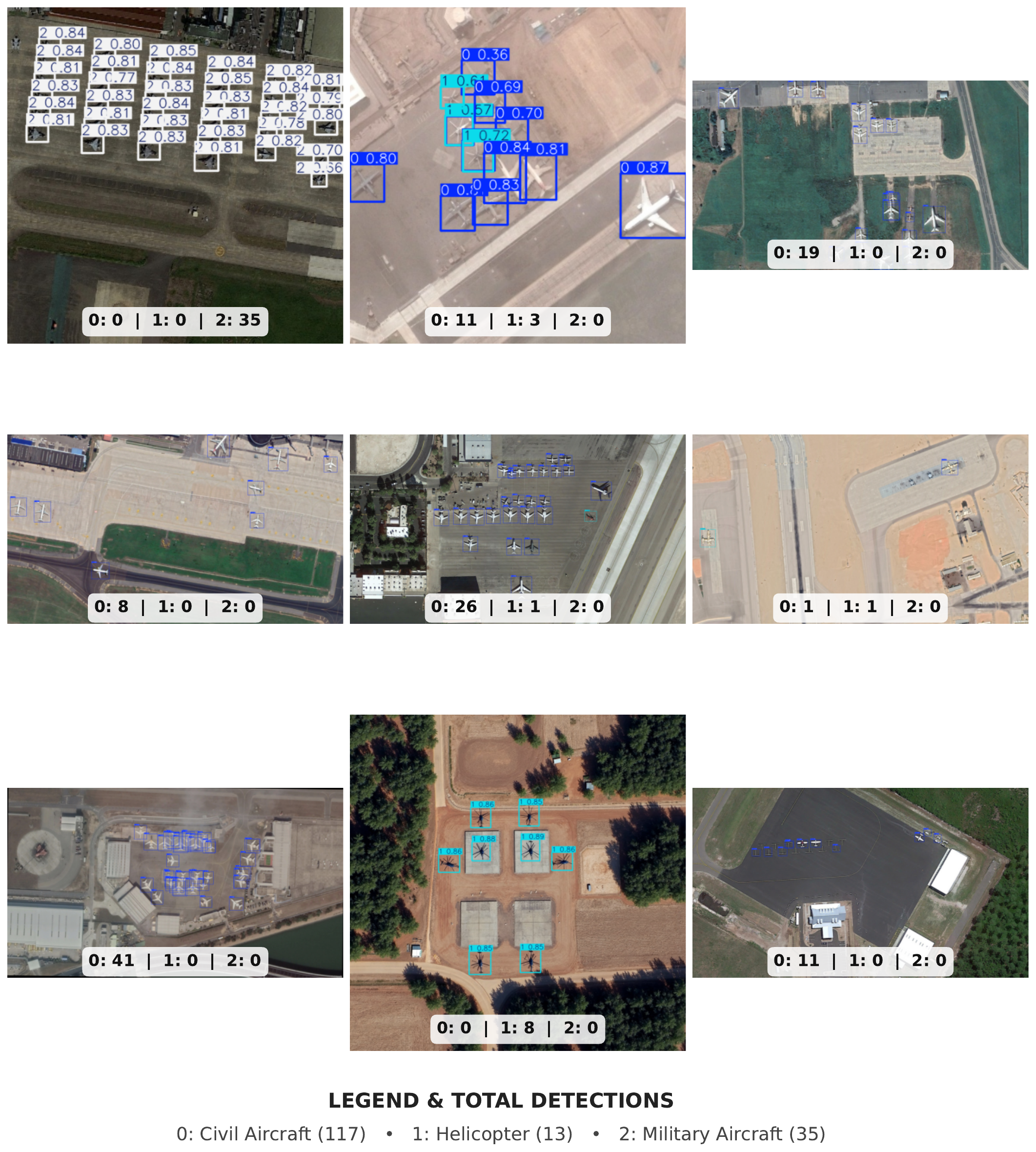}
    \caption{Inference mosaic on an airbase scene, illustrating the
    operational value of the pipeline for \gls{ISR} and \gls{GEOINT}
    applications.}
    \label{fig:isr_mosaic}
\end{figure}

As a proof of concept of the extensibility of the pipeline, the same
YOLO11n architecture was retrained on an extended 22-class
dataset comprising specific aircraft models (F-16, F-22, B-52, U-2,
C-130, E-3, \dots) plus the \textit{Runway} class. Evaluated on a
strictly isolated test partition to rule out overfitting, the resulting
normalised confusion matrix (Figure~\ref{fig:cm_22class}) shows a
diagonal above 95\,\% for almost every class. This high accuracy
illustrates the feature extraction of the augmented dataset rather 
than mere data memorisation. The only exception is \textit{Runway}, 
whose accuracy drops to 37\,\% due to confusion with linear background 
structures a limitation that motivates the future migration to 
instance segmentation discussed in Section~\ref{sec:conclusions}. An 
illustrative inference mosaic on an airbase scene is shown in 
Figure~\ref{fig:isr_mosaic}, illustrating the operational value of 
the pipeline for ISR and GEOINT applications: the on-board detector 
localises and classifies aircraft across the captured area, and only 
the confirmed metadata would be downlinked. 

\begin{figure}[h!]
    \centering
    \includegraphics[width=0.8\linewidth]{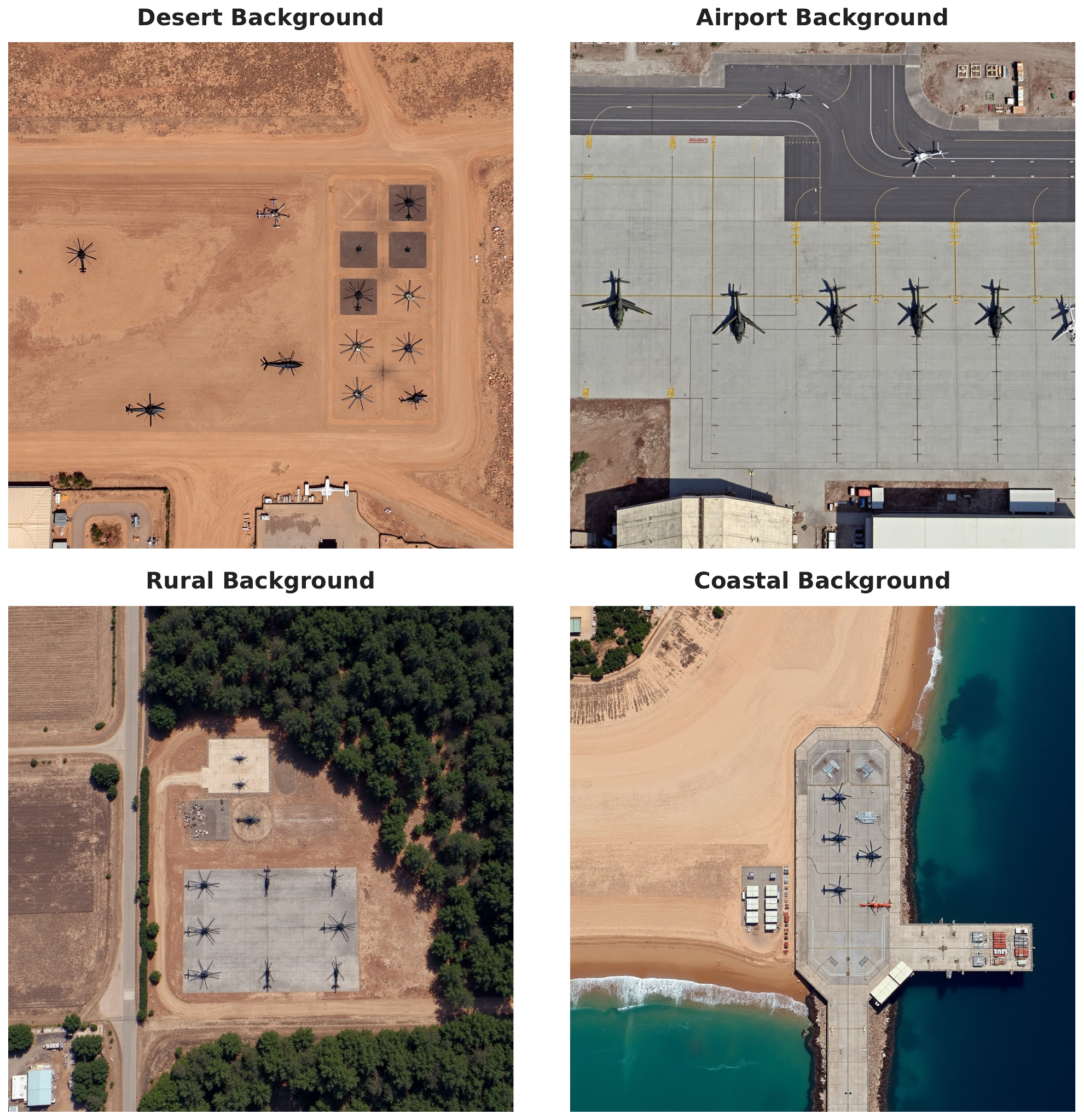}
    \caption{Synthetic helicopter imagery generated by the
    \gls{FLUX}+\gls{LoRA} pipeline across the four operational
    environments.}
    \label{fig:synth_heli}
\end{figure}

Representative synthetic helicopter samples generated by the
\gls{FLUX}+\gls{LoRA} pipeline are shown in
Figure~\ref{fig:synth_heli}: the photorealistic fidelity and the
diversity of backgrounds support the empirical gains reported above.

\section{Discussion}\label{sec:discussion}

This section interprets the empirical findings presented in Section~\ref{sec:results}, contextualising the performance of the proposed YOLO11n pipeline within the broader framework of edge computing for nanosatellites. Having demonstrated that the combination of a SWaP-aware detector and generative data augmentation successfully overcomes severe class imbalance, we now evaluate the real-world implications of these results. Section~\ref{sec:disc_applicability} explores the operational impact of this autonomous architecture for ISR and GEOINT missions, detailing its strategic advantages over traditional bent-pipe paradigms. Finally, Section~\ref{sec:disc_limitations} critically examines the boundaries of the current study, identifying methodological constraints and outlining the necessary steps toward physical hardware validation.

\subsection{Practical applicability in ISR and GEOINT}\label{sec:disc_applicability}

The experimental results support the proposed workflow as a
decision-support tool for real-time, autonomous airborne surveillance
from nanosatellites. By moving inference on board the 6U \gls{CubeSat},
the platform evolves from a passive data collector, characteristic of
the conventional bent-pipe architecture, into an autonomous node able
to filter redundant telemetry and deliver only tactical intelligence.
In an \gls{ISR} or \gls{GEOINT} context, this translates into two
operational advantages. First, the on-orbit inventory of airbases
enabled by the extended 22-class model (Section~\ref{sec:results_extended})
allows automated order-of-battle estimation without downlinking raw
imagery. Second, the intelligent-transmission ConOps of
Section~\ref{sec:platform_conops} reduces the decision latency from
hours or days, typical of ground post-processing, to fractions of a
second, which is critical for tracking dynamic targets such as
departing or manoeuvring aircraft. The trade-off between the
conventional and the proposed architectures, quantified in
Table~\ref{tab:edge_vs_bentpipe}, shows that the energy spent by the
Edge \gls{TPU} is largely offset by the massive savings in
radio-frequency transmission, making the approach energetically
sustainable within the 10--30~W budget of a \gls{CubeSat}.

\subsection{Limitations}\label{sec:disc_limitations}

Despite these promising outcomes, the proposed approach still faces a
number of limitations that should be addressed before an operational
deployment.

\begin{itemize}
  \item \textbf{Software-only validation.} All \gls{FPS} figures
  reported in Section~\ref{sec:results_deployment} are throughput
  projections obtained by scaling the measured T4 latency with the
  throughput-to-power ratio of the Edge \gls{TPU}; no
  Hardware-in-the-Loop test on a real flight-grade accelerator, nor
  radiation tests (\gls{SEE}/\gls{TID}), have been conducted yet. The
  physical deployment of the compiled model on an Edge \gls{TPU} or a
  Jetson Orin Nano, with measured power and per-frame latency under
  vacuum and radiation, is left as future work.
  \item \textbf{Runway-class performance.} In the extended 22-class
  model, the \textit{Runway} class drops to 37\,\% accuracy due to
  confusion with linear background structures. Bounding-box detection
  is not the most adequate formulation for elongated infrastructure,
  and a migration to instance segmentation (e.g.\ \textit{YOLO-seg})
  is expected to address this.
  \item \textbf{Convergence regime.} The balanced model was trained
  for only 20 epochs; although the results are already competitive, a
  longer convergence regime has not been characterised, and the
  diminishing-returns region remains open.
  \item \textbf{Daylight and clear-weather dependence.} The optical
  payload constrains the operational envelope to daylight and
  cloud-free scenes. Night and heavy-cloud operativity would require
  \gls{SAR} or \gls{IR} sensors, which are out of the scope of this
  work.
  \item \textbf{Generative background bias.} The \gls{FLUX}+\gls{LoRA}
  pipeline is sensitive to the prompt and the trigger-token
  engineering; the model should be re-validated whenever the area of
  interest is changed, to prevent background bias from injecting
  artefacts into the synthetic distribution.
\end{itemize}

These limitations are consistent with those reported in the broader
edge-AI literature for space~\cite{mystkowska2025hardware,
lentaris2023performance, chanoui2025trends} and do not invalidate the
core contribution of the paper, which is to demonstrate the
feasibility of an integral workflow that jointly addresses on-board
inference, generative data augmentation and \gls{SWAP}-aware
architecture selection for airborne surveillance from nanosatellites.
\section{Conclusions and Future Work}\label{sec:conclusions}

This paper has proposed and validated a workflow oriented to the
deployment of computer-vision algorithms on board nanosatellites,
showing the algorithmic and systemic feasibility of operating
under the \textit{edge computing} paradigm. The communication
bottleneck of conventional bent-pipe architectures has been mitigated
by moving inference on board a 6U \gls{CubeSat} equipped with a
Google Coral Edge \gls{TPU}, so that only confirmed detection
metadata are downlinked. A \gls{SWAP}-aware benchmark of three
one-stage detectors has been carried out, and \gls{YOLO}11n has been
selected as the architecture that provides the best balance between
precision and computational lightness, being the only candidate whose
\gls{INT8}-quantised weight (approximately 2.5~MB) fits the 8~MB
on-die \gls{SRAM} of the accelerator. The deficit of the minority
class in the original \textit{HRPlanesV2} dataset has been resolved
with a generative data-augmentation pipeline based on \gls{FLUX} and
\gls{LoRA}: the synthetic, pseudo-labelled helicopter imagery has
balanced the dataset to more than 53\,000 instances and has lifted
the helicopter F1 from 0.683 (classical augmentation alone) to 0.811,
raising the global mAP@50 from 77.9\,\% to 82.2\,\%. Finally, the
technical viability of the deployment on commercial hardware
compatible with the \gls{SWAP} budget of a \gls{CubeSat} has been
confirmed, with a projected on-orbit throughput of 25--30~\gls{FPS}
through \gls{SAHI} inference of $1024\times1024$ patches.

Despite these promising outcomes, the proposed approach still faces
limitations, especially regarding the software-only validation of the
on-board accelerator and the residual difficulty in detecting
elongated infrastructure such as runways. The challenge lies in
reconciling the dual objective of maximising detection precision and
ensuring deployability under the strict \gls{SWAP} constraints of a
nanosatellite, a problem that intensifies as the diversity of target
classes widens.

Future work could explore several lines of research. First, the
integration of \gls{SAR} and \gls{IR} sensors would enable 24/7
operativity regardless of meteorological conditions or the hour of
the day. Second, moving from static image detection to on-board video
tracking through algorithms such as ByteTrack or BoT-SORT, coupled
with the already trained \gls{YOLO}11n, would allow the estimation of
kinematic parameters such as speed, heading and predicted trajectory.
Third, the lightness of the nano architecture makes it exportable to
\glspl{UAV} as alternative host platforms, enabling real-time threat
identification on board reconnaissance drones without downlinking the
video to a command base. Fourth, the \gls{FLUX}+\gls{LoRA} pipeline
could be redirected towards counter-camouflage training, forcing the
detector to learn hidden patterns when an adversary attempts to
conceal assets. Finally, the physical deployment of the compiled model
on a space-grade accelerator --- an Edge \gls{TPU} or a Jetson Orin
Nano --- with measured power consumption and per-frame latency under
vacuum and radiation, would provide the definitive validation of the
concept and is the natural continuation of this work.

\section*{Declaration of competing interest}
The authors declare that they have no known competing financial
interests or personal relationships that could have appeared to
influence the work reported in this paper.

\section*{Data availability}
\ifanonymous
\else
The source code, trained weights, FLUX LoRA adapters and ComfyUI workflows supporting this study's findings are publicly available in the following repository to ensure that the results are reproducible: \url{https://github.com/Antonio23013/TFG-DETECCION-DE-OBJETOS.git}.
\fi

\section*{CRediT authorship contribution statement}
 \textbf{Antonio Delgado-Rosa:} Writing – original draft, Visualisation, Validation, Software, Methodology, Data curation, Conceptualisation. \textbf{David Muñoz-Valero:} Conceptualization, Supervision, Methodology, Validation, Writing -- review \& editing. \textbf{Enrique Adrian Villarrubia-Martin:} Writing – review \& editing, Data curation, Formal analysis. \textbf{Juan Moreno-García:} Conceptualization, Supervision, Formal analysis, Writing -- review \& editing, Funding acquisition.

\ifanonymous
\else
\section*{Acknowledgments}
This work was supported by grant PID2025-168152NB-C32 funded
by MCIN/AEI/10.13039/501100011033, by ERDF A Way of Making Europe. It was completed when Enrique Adrian Villarrubia-Martin was a predoctoral fellow at the Universidad de Castilla-La Mancha funded by the European Social Fund Plus (ESF+).
\fi

\bibliographystyle{elsarticle-num-names}
\bibliography{refs}


\end{document}